\documentclass[letterpaper, 10 pt, journal, twoside]{ieeetran} % ral

\usepackage{times}
\usepackage{epsfig}
\usepackage{graphicx}
\usepackage{amsmath}
\usepackage{amssymb}
\usepackage{algorithm}
\usepackage[noend]{algpseudocode}
\usepackage{subfigure}
\usepackage{color}

\IEEEoverridecommandlockouts                              % This command is only needed if 
\usepackage{fancyhdr}

\title{% \LARGE \bf %- ral
Ambiguity in Sequential Data: Predicting Uncertain Futures with Recurrent Models
}

\author{Alessandro Berlati$^{1, *}$, Oliver Scheel$^{2, 3, *}$, Luigi Di Stefano$^{1}$, Federico Tombari$^{3, 4}$% <-this % stops a space
\thanks{* indicates equal contribution}% <-this % stops a space
\thanks{$^{1}$Department of Computer Science and Engineering, Universit\`{a} di Bologna, 
Bologna, Italy}%
\thanks{$^{2}$BMW Group, M\"{u}nchen, Germany
        {\tt\small oliver.scheel@bmw.de}}%
 \thanks{$^{3}$Faculty of Computer Science, Technische Universit\"{a}t M\"{u}nchen,
Garching bei M\"{u}nchen, Germany}%   
\thanks{$^{4}$Google, Z\"{u}rich, Switzerland}%  
}

\begin{document}

\setlength{\textfloatsep}{5pt plus 1.0pt minus 2.3pt}

\maketitle
\thispagestyle{fancy} % - ral
%\pagestyle{empty}

%%%%%%%%%%%%%%%%%%%%%%%%%%%%%%%%%%%%%%%%%%%%%%%%%%%%%%%%%%%%%%%%%%%%%%%%%%%%%%%%
\begin{abstract}

Ambiguity is inherently present in many machine learning tasks, but especially for sequential models seldom accounted for, as most only output a single prediction.
In this work we propose an extension of the Multiple Hypothesis Prediction (MHP) model to handle ambiguous predictions with sequential data, which is of special importance, as often multiple futures are equally likely. Our approach can be applied to the most common recurrent architectures and can be used with any loss function. 
Additionally, we introduce a novel metric for ambiguous problems, which is better suited to account for uncertainties and coincides with our intuitive understanding of correctness in the presence of multiple labels. We test our method on several experiments and across diverse tasks dealing with time series data, such as trajectory forecasting and maneuver prediction, achieving promising results. 

\end{abstract}

% \begin{IEEEkeywords} Deep Learning in Robotics and Automation; Autonomous Agents; Intelligent Transportation Systems \end{IEEEkeywords}

%%%%%%%%%%%%%%%%%%%%%%%%%%%%%%%%%%%%%%%%%%%%%%%%%%%%%%%%%%%%%%%%%%%%%%%%%%%%%%%%

\section{Introduction}
\begin{figure}[!t]
\centering
\includegraphics[scale=0.3]{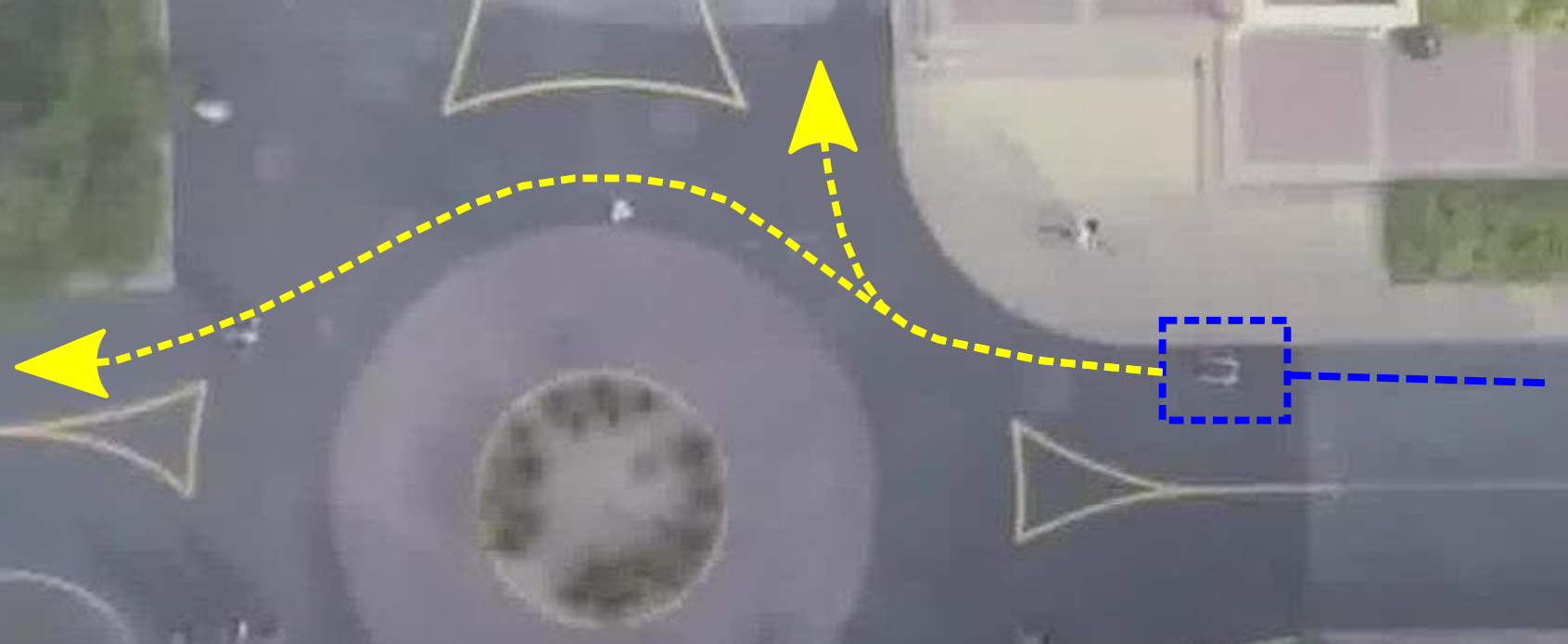}
\caption{When arriving at a roundabout, the cyclist marked in blue has multiple possibilities of continuing its path. At time of recording, a most likely one cannot yet be determined. A good prediction model should account for all possible trajectories at this point in time, and later converge to one, as soon as the cyclist's destination becomes clear.}
\label{fig:teaser}
\end{figure}

Ambiguity and uncertainty are inherently present in many machine learning tasks, both of sequential and non-sequential nature. A vehicle approaching an intersection might turn left or right, while in text generation (used, e.g.,  in mobile phones for auto-completion) multiple characters or words might be equally likely to follow the current one. Figure \ref{fig:teaser} visualizes the multiple trajectories possible when encountering a roundabout.
%In this work we propose an extension of the Ambiguous Prediction Model for classical, non-sequential models introduced by Ruprecht et al. \cite{Rupprecht2017LearningIA} to recurrent models. We cover the three most prominent and successful architectures for sequential models, thus providing a general framework for a wide range of problems. Additionally, we introduce a new metric tailored for ambiguous problems, as we found previous ones to be unsuited. Many previous works use some form of oracle metric, which only considers the best hypothesis \cite{desire, socialgan}. Although theoretically sound, this could easily be fooled by guessing solutions.

%Further, calculating the average over multiple predictions does not coincide with our intuitive understanding of ambiguity, consider Figure \todo{figure}.
%: When considering the problem of predicting trajectories at intersections, the output of a standard single hypotheses model compares to the "better" output of a multiple hypotheses model, compare Figure X \todo{Make figure. Show intersection. Can also introduce the problem here as well as the metric}

Several papers have addressed problems stemming from ambiguities and uncertainties, although, mainly, using problem-specific solutions 
% which cannot be transferred to other tasks 
\cite{socialgan, desire, 7780620}. One general solution are Mixture Density Networks (MDNs) \cite{mdn}. However, they are  defined only for regression problems and have some practical limitations \cite{Rupprecht2017LearningIA}. Another line of work concerns Multiple Choice Learning \cite{Lee2016StochasticMC}, in which ensembles of $M$ models are used. This uses more parameters and does not share information between predictions. Furthermore, most works in this field focus on non-sequential problems like image classification tasks, and pay little to no attention to time series data.

In this work we propose an extension to recurrent architectures of the Multiple Hypothesis Prediction (MHP) model by Rupprecht et al. \cite{Rupprecht2017LearningIA}, originally devised for feed-forward, non-sequential architectures. 
It is a general multi-purpose ambiguous prediction framework, specifically designed for sequential and recurrent models, and can be used to turn any existing deterministic model into one outputting multiple predictions.
%To the best of our knowledge, ours is the most general multi-purpose ambiguous prediction framework specifically designed for sequential data and recurrent models.
% In this work we introduce the first multi-purpose ambiguous prediction framework for recurrent models, based on \cite{Rupprecht2017LearningIA}. 
%Instead of explicitly modelling a probability distribution over the output, as it is done by MDNs, 
Core idea is using an implicit Voronoi tessellation of the label space, while any arbitrary loss function can be used.
%The $M$ predictions are obtained in a more sophisticated way than duplicating whole models, as it is done in Multiple Choice Learning \cite{Lee2016StochasticMC}, thereby saving parameters and enabling information exchange between hypotheses.  
By considering a sequence-to-sequence prediction model, an encoder-decoder architecture and a model for generating sequences, we cover the most important recurrent models as of today, showing for each how to incorporate ambiguity.
We test our models on multiple problems and find that they outperform standard non-ambiguous models on each of them and compare favorably against related methods.

Additionally, we introduce a new metric tailored for ambiguous problems, as we found previous ones to be unsuited. Indeed, many previous works use some form of oracle metric, which only considers the best hypothesis \cite{socialgan, desire}. Although theoretically sound, this could easily be fooled by guessing a multitude of diverse solutions in hopes of approximating the correct one. Conversely,
the proposed metric better captures our understanding of ambiguity. This is achieved by re-labelling data samples, possibly assigning multiple labels to single data points, and requiring predictions to match all of them. 
\section{Related Work}
\label{sec:related}
In last years deep neural networks, most notably Convolutional Neural Networks (CNNs), have proven to be general and flexible function approximators.
% capable of outperforming other methods, and even humans, in a variety of tasks, like image classification \cite{imagenetoutperform}.
However, most works focus on learning a one-to-one mapping from input to output, thus optimizing models to predict the single best hypothesis. Although in ambiguous situations this is neither desired nor correct, prediction of multiple hypotheses has been addressed less frequently in research.
Rupprecht et al. introduced a general ambiguous prediction framework for feed-forward models dubbed MHP (Multiple  Hypothesis  Prediction), which is based on a Voronoi tessellation of the label space \cite{Rupprecht2017LearningIA}. Here we extend this principle to recurrent models for sequential data.
%\vspace{-\baselineskip}
\paragraph{General Models} \emph{Mixture Density Networks (MDNs)} are a general method for handling ambiguities \cite{mdn} based on learning the parameters of a Gaussian Mixture Distribution. However, in contrast to the previously mentioned ambiguous prediction framework, they are limited to regression problems, and model the encountered distributions explicitly instead of implicitly via the Voronoi tessellation.
Further, it was shown that MDNs can be difficult to train due to numerical instabilities in high dimensional spaces \cite{Rupprecht2017LearningIA}. \\
\emph{Multiple Choice Learning} focuses on training an ensemble of $M$ models to output $M$ predictions, backpropagating gradients only to the lowest error predictors  \cite{Lee2016StochasticMC}.
%Whereas early versions required costly retraining \cite{GuzmnRivera2012MultipleCL}, Lee et al. improved on this so to enable  usage of standard Stochastic Gradient descent  by backpropagating gradients only to the lowest error predictors for each example \cite{Lee2016StochasticMC}. 
This makes their minimum formulation similar to that of Rupprecht et al. \cite{Rupprecht2017LearningIA} and ours. 
%Through the MHP framework Lee's idea is extended by mathematical insights as to why this formulation works, and also to regression problems. 
Indeed, the MHP  framework provides mathematical insights on why their formulation works and enables to extend it to regression problems.
Further, instead of training $M$ different networks, a joint architecture is applied in \cite{Rupprecht2017LearningIA}, thus saving parameters and enabling information exchange. Although Lee et al. show how to address also the sequential problem of image caption generation by deploying  LSTMs \cite{Lee2016StochasticMC}, our approach differs from their work as it is  focused on sequential models and describes extensions of different state-of-the art architectures rather than simply copying the network $M$ times. \\
\emph{(Conditional) Variational Autoencoders} (CVAEs) are another common method for modelling uncertainty \cite{Kingma2013AutoEncodingVB}. This is done by learning a data-conditional latent distribution, s.t. diverse samples can be generated by sampling from this. Gregor et al. introduced sequential VAEs \cite{gregor2015draw}. 
%Jain et al. applied a similar principle to generate questions from images \cite{creativity}. 
Although the sequential VAE can be applied in principle to arbitrary problems, still it is a specific  type of network model, and existing works focus on problem-specific solutions. In contrast to this, our approach can be used to extend any sequential model to predict  multiple hypotheses. Furthermore, in VAEs often a Gaussian distribution is used as prior for the latent distribution, thus limiting the space of possible solutions, whereas our approach is able to  model implicitly arbitrary multi-modal distributions.
%\vspace{-\baselineskip}
\paragraph{Image Classification}  Handling multiple predictions has been addressed in image classification \cite{Gong2014DeepCR} as multiple labels may often be assigned to an image. Wang et al. combine a CNN with a Recurrent Neural Network (RNN) to predict multiple labels \cite{7780620}. Yet,  the solutions developed for image classification do address only non-sequential problems.
%\vspace{-\baselineskip}
\paragraph{Sequential Models} As for sequential problems, ambiguity has been considered less comprehensively. Indeed, most proposals deal with problem-specific approaches, typically leveraging on Generative Adversarial Networks (GANs), VAEs or Reinforcement Learning,
% possible_rebuttal_cut
or other kinds of sequential generative modelling \cite{naomi} in order to produce diverse outputs.
To predict trajectories, Lee et al. use RNNs combined with CVAEs \cite{desire} whilst Gupta et al. GANs and a special diversity loss \cite{socialgan}.
%Then, Shi et al. employ Inverse RL to avoid mode collapse when generating text \cite{textirl}, while Shao et al. stochastic beam search with sequence-to-sequence models \cite{beamsearch}.
Bazzani et al. apply MDNs to the output of RNNs at each timestep (called Recurrent MDNs), to obtain a saliency map of visual attention in videos \cite{bazzani2016recurrent}.
Unlike these problem-specific solutions, in this paper we propose a general framework to handle ambiguity in sequential problems. 
%\vspace{-\baselineskip}
\paragraph{Ambiguous Labelling} 
%Several works have addressed the topic of dealing with multiple labels. 
Kalyan et al. employ multiple labels in order to deal with sparse annotations \cite{Kalyan2018LearnFY}. %They learn jointly the model and an embedding function which defines similarity, i.e. the concept of neighborhood in the input space. 
%This allows modifying the loss function in order to take labels of neighbouring data points into account. On the one hand, this resembles our minimum formulation, although Kalyan et al. explicitly calculate distances in the label space and re-label data points, whereas our approach implicitly learns the underlying ambiguity, thereby addressing ambiguous predictions rather than ambiguous annotations. On the other hand, t
The underlying principle stems from the same motivation as our newly proposed metric, namely clustering points w.r.t. the label space and reassigning labels. However, while theirs is learned in combination with the model, ours is a fixed deterministic mapping intended to be used as a problem and model independent metric. 
% Likewise, other papers dealing with multiple labels within model formulation \cite{Gao:2017:DLD:3083729.3101479, 6909633} do not propose any general-purpose metric to handle ambiguity.   
Rhinehart  et al. introduce two entropy terms as loss functions for trajectory prediction, encouraging predictions to be diverse whilst simultaneously precise \cite{rhinehart2018r2p2}. While again very similar in motivation, this also is no metric. Further we introduce ours in a more general way, e.g. also discussing classification problems, and describing how arbitrary metrics can be extended in our way, s.t. well-known metrics like Precision and Recall can be used and compared in a multi-modal setting. 
%We further feel that the relabelling step allows for better interpretability.

%address ambiguous labeling withinh have addressed the problem of dealing with multiple labels, though mostly

\section{Methodology}
\subsection{Prerequisites}
In this section we introduce notation and summarize the MHP Model \cite{Rupprecht2017LearningIA}.
\subsubsection{Multiple Hypothesis Prediction Model}
Let $\mathcal{X}$ and $\mathcal{Y}$ be vector spaces of input and output variables, or labels, of a problem, and $N$  the number of available data samples. Further, let $p(x, y)$ denote the joint probability distribution over input variables and labels.

% In a classical supervised setting, in which samples ($x_i, y_i$) are drawn from $p(x, y)$, we are interested in training a predictor $f_{\theta}: \mathcal{X} \rightarrow \mathcal{Y}$, which minimizes the error
% \begin{equation}
%     \frac{1}{N}\sum_{i=1}^N \mathcal{L}(f_{\theta}(x_i), y_i),
% \label{eq:error}
% \end{equation}
% where $\mathcal{L}$ is an arbitrary loss function and the predictor is characterized by its parameters $\theta$.
% Equation (\ref{eq:error}) is an approximation of the continuous formulation
% \begin{equation}
%     \int_{\mathcal{X}} \int_{\mathcal{Y}} \mathcal{L}(f_{\theta}(x), y)p(x, y) \ dy \ dx.
%     \label{eq:contloss}
% \end{equation}
% which is minimized by the conditional average
% \begin{equation}
%     f_{\theta}(x) = \int_Y y \cdot p(y|x) \ dy.
% \end{equation}

Whereas in a classical supervised setting we are interested in training a predictor $f: \mathcal{X} \rightarrow \mathcal{Y}$, in the MHP Model \cite{Rupprecht2017LearningIA} the prediction function is extended with the possibility of outputting $M$ predictions, or hypotheses:
\begin{equation}
    f(x) = (f^1(x), \ldots, f^M(x)).
\end{equation}
When computing the loss $\mathcal{L}$, the best among the $M$ predictors is considered, resulting in the continuous formulation
\begin{equation}
    \int_{\mathcal{X}} \sum_{j=1}^M \int_{\mathcal{Y}_j(x)} \mathcal{L}(f^j(x), y)p(x, y) \ dy \ dx.
\label{eq:mhploss}
\end{equation}
Here, $\mathcal{Y} = \cup_{i=1}^M \mathcal{Y}_i$ is the Voronoi tessellation of the label space, which is induced by $M$ generators $g^j(x)$ and the loss $\mathcal{L}$:
\begin{equation}
    \mathcal{Y}_j(x) = \{y \in \mathcal{Y}: \mathcal{L}(g^j(x), y) < \mathcal{L}(g^k(x), y) \ \forall k \neq j\}.
\end{equation}
\cite{Rupprecht2017LearningIA} proves that, in order for Equation \ref{eq:mhploss} to be minimal, the generators have to equal the $M$ predictors and that $f^j$ predicts the conditional mean of the Voronoi cell it defines.
%To avoid confusion, from now on we drop the subscript $\theta$ denoting the model parameters.
A full minimization scheme defined from scratch can be found in \cite{Rupprecht2017LearningIA}. 
% however, it can be easily implemented using a meta-loss on top of any loss $\mathcal{L}$:
Yet, as shown in \cite{Rupprecht2017LearningIA}, the method can be implemented efficiently by using a meta-loss on top of any loss $\mathcal{L}$:
\begin{equation}
    M(f(x_i), y_i) = \sum_{j=1}^M \hat{\delta}(y_i \in \mathcal{Y}_j(x_i))\mathcal{L}(f^j(x_i), y_i).
\end{equation}
% The Kronecker delta $\delta$ returns 1, if its enclosed condition is true, otherwise 0, and is used to select the best hypotheses. In practise we have to relax this condition, as all predictors could be initialized so far from the target, that all hypotheses lie in a single Voronoi cell, making this the only one to be updated.
% Thus, a relaxation using a weight $\epsilon$ ($0 < \epsilon < 1$) is used:
% \begin{equation}
%      \hat{\delta}(a)=\begin{cases}
%     1 - \epsilon, & \text{if a is true},\\
%     \frac{\epsilon}{M-1}, & \text{otherwise}.
%   \end{cases}
% \end{equation}
where the Kronecker delta $\hat{\delta}$ is defined by
\begin{equation}
     \hat{\delta}(a)=\begin{cases}
    1 - \epsilon, & \text{if $a$ is true},\\
    \frac{\epsilon}{M-1}, & \text{otherwise},
  \end{cases}
\end{equation} and used to select the best hypotheses.
The $\epsilon$ relaxation is used, as a bad initialization could place all hypotheses in a single Voronoi cell.
Note that $M$ is a hyper-parameter, which can be chosen freely, and that most models in this field require such a hand-tuned model parameter \cite{Lee2016StochasticMC, mdn}.
\subsubsection{Long Short-Term Memory Cells}
The main goal of this work is to extend the above mentioned prediction method to a variety of different recurrent models. For all of these, we use recurrent neural networks consisting of Long Short-Term Memory Cells (LSTMs) \cite{LSTM}. We refer the reader to \cite{LSTM} for a comprehensive  description of LSTMs. Here, for notational purposes, we bundle all internal calculations in a function referred to as $\texttt{LSTM}$, such that one computation step can be written as
\begin{equation}
    (\mathbf{h}_t, \mathbf{\tilde{c}}_t) = \texttt{LSTM}(x_t, \mathbf{h}_{t-1}, \mathbf{\tilde{c}}_{t-1})
\end{equation}
with $\mathbf{x}_t$ denoting the input at timestep $t$, $\mathbf{h}_t$ and $\mathbf{\tilde{c}}_t$ the hidden state and cell state at time $t$.
\subsection{Sequence-to-Sequence Prediction}
\label{sec:s2s}
\begin{figure}[!t]
\centering
\includegraphics[scale=0.5]{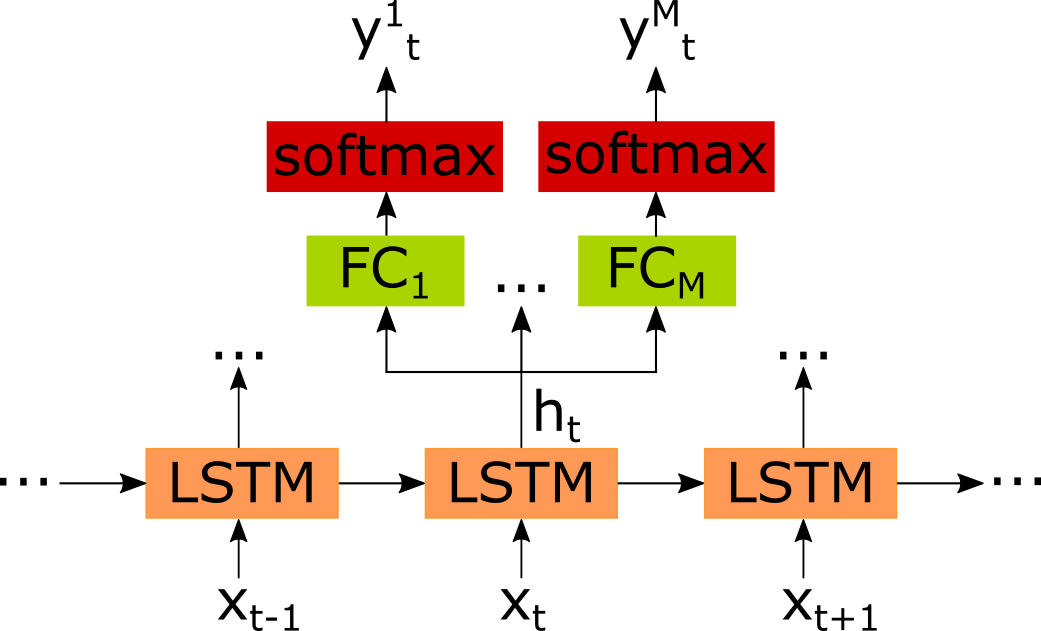}
\caption{Visualization of the sequence-to-sequence model. $FC_m$ denotes a fully connected layer with parameters $W_m$ and $b_m$.}
\label{fig:seq2seq}
\end{figure}

Our first proposed model is an extension of the classical sequence-to-sequence prediction architecture. In this, we are given an input sequence and expected to return a prediction in each step, thus also returning a sequence.
% possible_rebuttal_cut
%Exemplary applications are prediction of driving maneuvers \cite{B4C}. % as well as, in case all frames have to be classified, human activity recognition \cite{ntu}. 
Thus, each sample $(x_i, y_i)$ is now made out of sequences $x_i = (x_{i, 1}, \ldots, x_{i, n})$, $y_i = (y_{i, 1}, \ldots, y_{i, n})$.
To obtain an MHP Model, we replace the common fully-connected layer on top of the recurrent network with $M$ copies of it  which do not share weights (see Figure \ref{fig:seq2seq}):
\begin{equation}
\label{eq:lstm}
\begin{split}
\mathbf{y}_t^1 &= \texttt{softmax}(\mathbf{W_1} \mathbf{h}_t + \mathbf{b_1})\\
\ldots \\
\mathbf{y}_t^M &= \texttt{softmax}(\mathbf{W_M} \mathbf{h}_t + \mathbf{b_M})\\
\end{split}
\end{equation}
Thus, the loss function needs to be applied to sequential data. In our tested problems we use the sequential cross-entropy loss as follows: If $\mathbf{\tilde{y}}_{i, t}^j$ are the predicted probabilities of the ground truth at time $t$, then for a single sample the standard softmax loss is calculated as $-log(\mathbf{\tilde{y}}_{i, t}^j)$, hence: 
\begin{equation}
\label{eq:crossentroy_sequential_loss}
\mathcal{L}(f^j(x_i), y_i) = \frac{1}{n}\sum_{t=1}^{n}-log(\mathbf{\tilde{y}}_{i, t}^j) 
\end{equation}
\subsection{Encoder-Decoder}
\label{sec:enc-dec}
\begin{figure}[!t]
\centering
\includegraphics[scale=0.5]{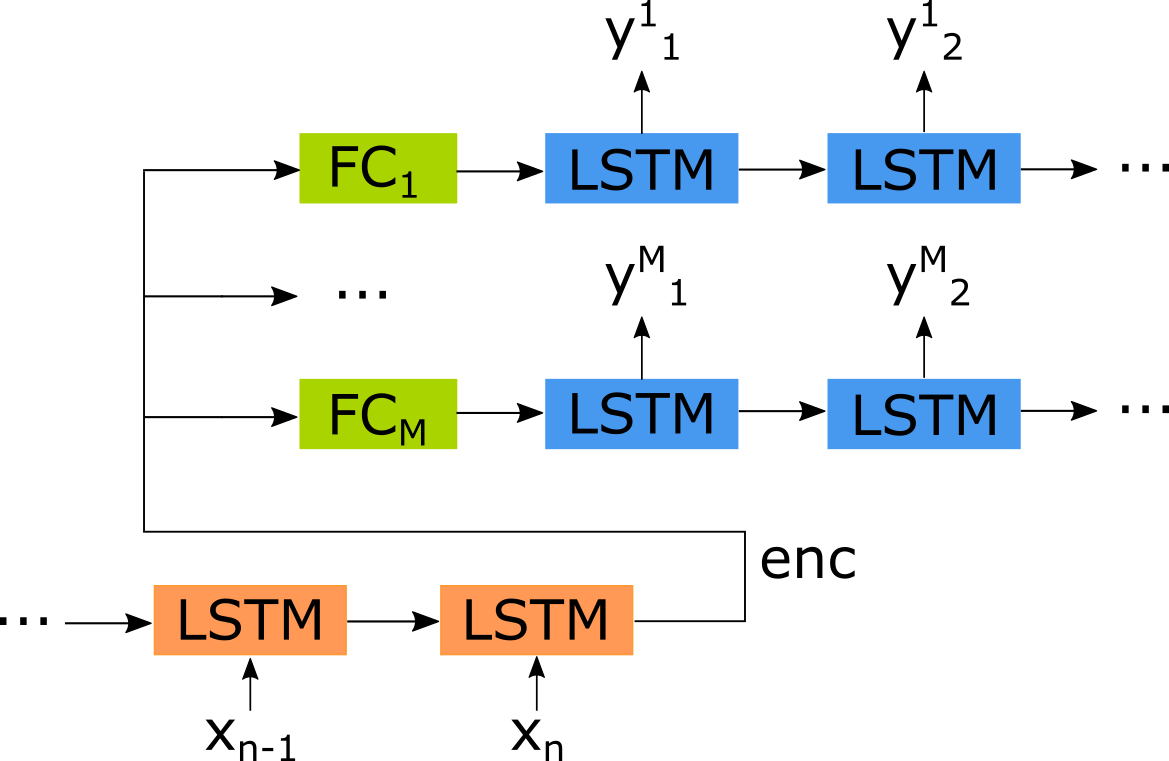}
\caption{Visualization of the encoder-decoder model. The encoder is drawn in orange, the decoder in blue. $FC_m$ denotes a fully connected layer with parameters $W_m$ and $b_m$.}
\label{fig:enc-dec}
\end{figure}

% possible_rebuttal_cut
Encoder-decoder architectures excel in many sequential problems like trajectory prediction \cite{socialgan, desire}. 
Encoder-decoder architectures consist of two separate recurrent networks: the encoder processes the input sequence $x_i = (x_{i, 1}, \ldots, x_{i, n})$ and eventually produces a high-dimensional vector representation $\mathbf{enc}$, which is then fed to the decoder to produce the output sequence $y_i = (y_{i, 1}, \ldots, y_{i, m})$. 
We extend this model to handle ambiguity by introducing a fully connected layer between the encoder and the decoder, so to produce $M$ vectors (see Figure \ref{fig:enc-dec}):
\begin{equation}
\label{eq:enc-dec}
\begin{split}
\mathbf{enc} &= \texttt{Encoder}(x_i)\\
\mathbf{y}^1_i &= \texttt{Decoder}(\mathbf{W_1} \mathbf{enc} + \mathbf{b_1})\\
\ldots \\
\mathbf{y}^M_i &= \texttt{Decoder}(\mathbf{W_M} \mathbf{enc} + \mathbf{b_M})
\end{split}
\end{equation}
As for the loss, any function suitable to sequences may be used. In our experiments dealing with encoder-decoder models we use the L2-loss $\mathcal{L}(f^j(x_i), y_i) = \frac{1}{m}\sum_{t=1}^{m}(\mathbf{y}^j_{i, t} - y_{i, t} )^2$.

\subsection{Sequence Generation}
\label{sec:gen}
Although for the problem of generating sequences the samples are still sequence pairs ($x_i, y_i$), the focus is to learn conditional probabilities of single frames $P(x_{i, t} \vert x_{i, t-1}, \ldots, x_{i, t-m})$. This way, after training, the network can be initialized with a seed and then run in a closed-loop, thereby creating sequences of arbitrary length. 
There is a wide range of possible applications, such as text generation \cite{DBLP:journals/corr/Graves13}. 
By using a similar model as in Section \ref{sec:s2s} (duplicating the output layer $M$ times), the training process remains the same, and the same loss functions may be used. 

During inference, though, to  employ an MHP model, we encounter the problem of exponentially growing possibilities in each step. This is not feasible, not just in terms of practical usage but also in terms of limited resource constraints. Therefore, during inference, our proposed solution makes use of different functions to decide when to split and when to merge predictions. These functions can be chosen in a problem-specific way, but it is also possible to train, e.g., a neural network for this task, implicitly calculating these functions. Note that this procedure is only applied during inference, s.t. these functions do not need to be differentiable. Similar issues have been known before, for related strategies (albeit employing SHP models regarding our definition) we refer to \cite{exp}.

Inference is started with a single hypothesis $H$, which can be empty, contain a special starting symbol or an arbitrary number of elements. In each step,  inferring for $d$ future steps is simulated. Due to the prediction of $M$ hypothesis in each of these, a tree of depth $d$ and branching factor $M$ is created. When the predictions in this tree are not yet diverse enough but close together, the first layer of the tree is merged into one prediction, which is appended to $H$. This merging is done by a function referred to here as \texttt{Merge}, while the function to check the tree diversity is denoted as  \texttt{CheckSplit}. When this encounters a diverse enough tree, function \texttt{ChooseTreePaths} finds $M$ hypotheses $h_1, \ldots, h_M$ of length $d$ in the tree. $H$ is subsequently split into $M$ different predictions by appending $h_1, \ldots, h_M$, and these hypothesis are consequently followed separately: For each hypothesis path the inference simulation is done, and the resulting $M$ predictions  are merged via \texttt{Merge}.
The pseudocode for this scheme is shown in Algorithm \ref{alg:seq}, while exemplar implementations of these black box functions are shown in Section \ref{sec:helpertoy}. %V-A.b.%\ref{sec:toypred}.
\begin{algorithm}
\scriptsize{
\caption{Inference for MHP Sequence Generation, starting with $start$ and inferring for a total of $l$ steps. $BuildTree$ expects the starting element as well as the desired tree depth as input.}
\label{alg:seq}
\begin{algorithmic}
\Procedure{\texttt{Infer}}{$start,d, l$}
\State \texttt{DoneSplit} = False;
\State \texttt{Predictions} = $start$; 
\For{i in 1 .. $l$}
\If{not \texttt{DoneSplit}}
\State p = last item of \texttt{Predictions}
\State \texttt{tree} = \texttt{BuildTree}(p, d);
\If {\texttt{CheckSplit}(\texttt{tree})}
\State $[(p_1^1, \ldots, p_{d}^1), \ldots, (p_1^M, \ldots, p_{d}^M)]$ = \texttt{ChooseTreePaths}(\texttt{tree}):
\State Append $([p_1^1, \ldots, p_1^M], \ldots, [p_d^1, \ldots, p_{d}^M])$ to \texttt{Predictions};
\State \texttt{DoneSplit} = true;
\Else
\State $[p^1, \ldots, p^M] = $ First layer of \texttt{tree};
\State   p = \texttt{Merge}($[p^1, \ldots, p^M]$);
\State Append p to \texttt{Predictions};
\EndIf
\Else
\State$[p_1, \ldots, p_M] = $ last item of \texttt{Prediction};
\For{m in 1 .. M}
\State p = \texttt{BuildTree}($p_m$, 1);
\State $p_{m}$ = \texttt{Merge(p)};
\EndFor
\State Append $[p_1, \ldots, p_M]$ to \texttt{Predictions};
\EndIf
\EndFor
\EndProcedure
\State \textbf{return} \texttt{Predictions}
\end{algorithmic}
}
\end{algorithm}

\section{Multi-Modal Metric}
\label{sec:mm-metrics}
In many works concerning multiple predictions, some form of oracle metric is used, which compares the ground truth to the closest prediction among possibly multiple ones \cite{desire, socialgan}. Formally, if $l$ is a metric between prediction set $X_i$ and label $y_i$, then $l_{oracle}=(X_i, y_i) = min_{x \in X_i}l(x, y_i)$. Although this rewards models for predictions close to the ``correct'' one and allows for diverse predictions, we argue that this metric should not be the gold standard for problems containing ambiguity. Indeed, better scores could be reached by simply guessing hypotheses, and in real-world settings it is not clear what actions should be taken based upon models with good oracle scores. Further, and even more importantly, ambiguous situations sometimes lead to multiple, equally ``correct'' predictions.
On the other hand, when averaging multiple predictions, in most cases the advantage of predicting multiple hypotheses cannot be shown numerically, as the loss is minimized by the (incorrect) mean.
% Consider the following sample (visualized in Figure \todo{X}): We have $n$ trajectories of car traversing an intersection, half of which take a left turn and the other hand a right turn. At position $X$ that direction is not yet known and a correct prediction should contain both these cases. A standard single hypotheses model will only predict one, and achieve an accuracy of $50 \%$. A multiple hypotheses model that predicts both would also score an accuracy of $0.5 * (0.5 * 1 + 0.5 * 0) + 0.5 * (0.5 * 0 + 0.5 * 1) = 0.5$ (both hypotheses are correct for $50 \%$ of all samples and wrong for the other.
Thus we propose a novel, multi-modal (M2) metric specifically designed for problems containing ambiguities and uncertainties. The core idea is to automatically re-label samples, possibly allowing multiple labels, and extending standard metrics $l(x_i, y_i)$ to a set-based calculation $l_{M2}(X_i, Y_i)$. Note that this principle can be used for basically all prediction tasks and metrics, i.e. also when dealing with non-sequential data. 
We derive our metric for both the discrete and continuous label cases.
\subsection{Discrete Labels}
First step concerns re-labelling of samples. Purposely, for each label class $c$ we calculate a polytope $P(c)$ in the input space, containing all points labeled as $c$ (denoted by $\lambda(x) = c$).
Then, for each sample $x_i$ with label $y_i$, the new label set is defined as $Y_i = \{y_i\} \cup \{c \mid c \in C, \ x_i \in P(c) \}$, where $C$ is the set of all labels. 
Such a polytope could be the convex hull. However, this does not scale to high dimensions and it is not reducible in size: especially for high-dimensional and complex data, outliers can inflate the convex hull, which is not desired.
Thus, we define the polytope $P(c) = \{x \mid \lambda(x) = c, \ min_{\tau} \le x \le max_{\tau} \}$, where $x$, $min_{\tau}$ and $max_{\tau}$ are vectors of the d-dimensional sample space. 
A threshold $\tau$ can be defined to control the size of the polytope, denoting that an (approximate) fraction $\tau$ of all points labeled $c$ are contained in the polytope.
For this, $min^d_{\tau}$ and $max^d_{\tau}$ are given by the $([1 - \frac{\tau}{2}] \cdot 100$)th and $(\frac{\tau}{2} \cdot 100)$th percentile, respectively, in each dimension $d$.
$\tau$ should be chosen such that it best resembles the ambiguity of the problem, e.g. overlaps with our intuitive understanding of multi-modality. Thus it is freely choosable, nevertheless the results obviously are reproducible and comparable given $\tau$.
Although this method is applicable for an arbitrary number of dimensions (consider treating MNIST \cite{lecun1998mnist} images as 784-dimensional vectors), for high-dimensional spaces it is possible and recommendable to use a lower-dimensional latent space (e.g. intermediate layers of common CNN architectures or autoencoders, similar to \cite{Kalyan2018LearnFY}) and define the polytope on this.
For a better understanding, in Section \ref{sec:results} the resulting polytopes of a toy task are shown.
Basically, any metric can be extended with this new label set in a natural way. Here we consider the widely-used  metrics Precision (Pr) and Recall (Re).
The extension is motivated from the context of information retrieval, i.e. a Precision of $1$ should only be reached if all predicted hypotheses are correct, analogously a Recall of $1$ should mean that all labels are predicted.
To satisfy this we define
\begin{equation}
pr(f(x_i), Y_i) = \frac{\vert f(x_i) \cap Y_i\vert}{\vert f(x_i) \vert}
\end{equation} and 
\begin{equation}
re(f(x_i), Y_i) = \frac{\vert f(x_i) \cap Y_i \vert}{\vert Y_i \vert}
\end{equation}
for each sample $i$ and ambiguous prediction $f(x_i)$ and
overall $Pr_{M2} = \frac{1}{N}\sum_{i=1}^N pr(f(x_i), Y_i)$, $Re_{M2} = \frac{1}{N}\sum_{i=1}^N re(f(x_i), Y_i)$.
\subsection{Continuous Labels}
Still assuming the set-based interpretation, in the case of continuous labels we require that each prediction is close to at least one label, and that for each label there is at least one close hypothesis.
However, with continuous labels we cannot directly define the polytope but first apply clustering in the label space to get discrete labels $C$.
Hence, analogously, we define $P(c)$ and $Y_i = \{\mu(c) \mid c \in C, \ x_i \in P(c) \}$, but here $\mu(c)$ denotes the center of cluster $c$ (see Section \ref{sec:results} for visualizations of such clusters).
We define the extension of a metric $l(x_i, y_i)$ by 
\begin{equation}
\begin{split}
        l_{M2}(x_i, Y_i) & = \frac{\sum_{x \in f(x_i)} min_{y \in Y_i} l(x, y)  }{\vert f(x_i) \vert + \vert Y_i \vert} \\
        & + \frac{\sum_{y \in Y_i} min_{x \in f(x_i)} l(x, y)  }{\vert f(x_i) \vert + \vert Y_i \vert}
\end{split}
\end{equation}

\section{Problems and Datasets}
In this section we introduce several problems coming from different fields, together with the models used to address them. As the general models are introduced in Sections \ref{sec:s2s} to \ref{sec:gen}, here we will just highlight problem specific adaptations (if any).

\subsection{Toy Intersection}
\label{sec:helpertoy}
\begin{figure}[!t]
\centering
\includegraphics[scale=0.1]{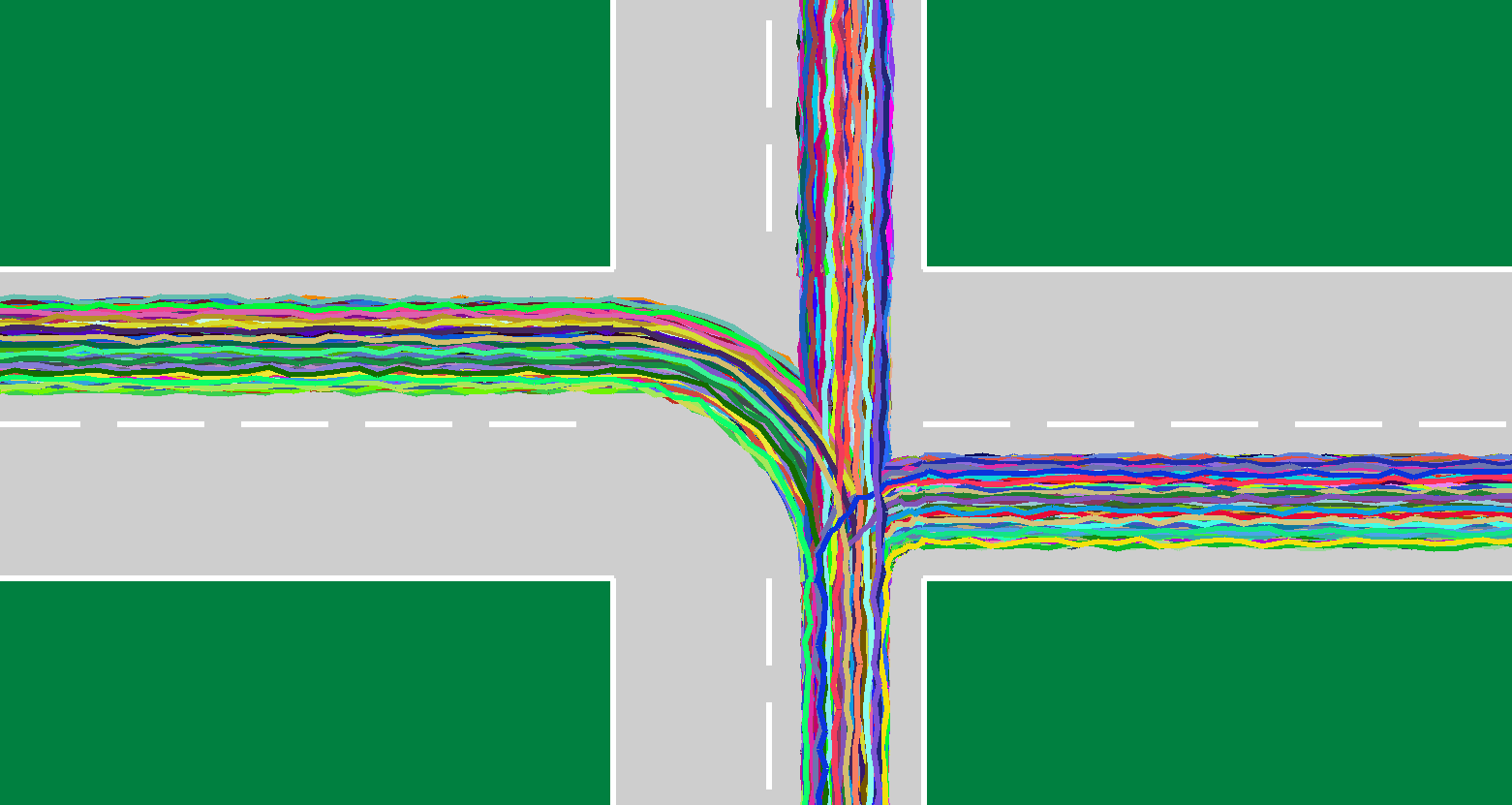}
\caption{Visualization of simulated trajectories on a three-way intersection. Each vehicle approaches the intersection from the bottom of the image, and then chooses one of three possible paths.}
\label{fig:trajectories}
\end{figure}

To motivate the usage of multiple hypotheses, we start with simple toy problems, which offer clear data. They are based on synthetically created traffic on a three-way intersection (see Figure \ref{fig:trajectories}). Each trajectory $t$ is an ordered list of 2D coordinates: $t = ([c_1^1, c_2^1], \ldots, [c_1^n, c_2^n])$.

\paragraph{Toy Classification}
\label{sec:toyclass}
Goal of the classification task is to predict each vehicle's destination (left, straight or right) at every time-step. We solve this problem by using the sequence-to-sequence prediction approach from Section \ref{sec:s2s}. Let $l_i$ be the correct label of sample $i$, then $x_i = ([c_1^1, c_2^1], \ldots, [c_1^n, c_2^n])$ and $y_i = (l_i^1, \ldots, l_i^n)$.

\paragraph{Toy Prediction}
\label{sec:toypred}
For this task future trajectories have to be predicted conditioned on past ones. We showcase the usage of the encoder-decoder and sequence generation models (Sections \ref{sec:enc-dec} and \ref{sec:gen}). For both models, a first part of each trajectory $x_i = ([c_1^1, c_2^1], \ldots, [c_1^m, c_2^m])$, $m < n$, serves as input, while $y_i = ([c_1^{m+1}, c_2^{m+1}], \ldots, [c_1^n, c_2^n])$ is the desired output. 
In the encoder-decoder model $x_i$ and $y_i$ are the input and output of the encoder and decoder, respectively. 

As for the sequence generation model, it is trained with full trajectories, learning the conditional probabilities of successive trajectory points. During inference, the model is initialized with $x_i$ and expected to produce $y_i$, constructing a tree of depth 8 in each step. Function \texttt{CheckSplit} checks for the maximum distance of two points in the last layer of the tree and returns true if this exceeds a certain threshold. \texttt{Merge} simply returns the mean of the predicted points. \texttt{ChooseTreePaths} sorts all points in the last layer of the tree by their angle to the current point, and then returns those $M$ points that evenly split the range of angles into $M$ parts.
\subsection{Lane Change Prediction}
\label{sec:LCP}
Predicting maneuvers, or more generally action anticipation, is a crucial task in many fields, e.g. autonomous driving, and has sparked great interest in the scientific community \cite{B4C, jiang2014modeling}.
Here we want to predict lane change maneuvers of vehicles as early and accurately as possible. We use the sequence-to-sequence prediction model from Section \ref{sec:s2s}.  At each timestep, five features $f_t$, like distance to the lane boundaries and lateral velocity, are used as inputs. Each timestep is assigned a label $l_t$, which can either be L (change left), F (follow lane), R (change right)  (see \cite{lcpattention} for more details). 
% A duration of three seconds before a lane change is labelled with the respective direction of the lane change, shortly before that and after the execution of the maneuver buffer labelswith weight 0 are added, the reamining time steps are labeled F. 
We use the publicly available NGSIM dataset \cite{ngsim}, which offers vehicle trajectories recorded on highways and interstates in the United States.

\subsection{Trajectory Prediction}
A multitude of models have been proposed for the problem of predicting trajectories, mainly concerning pedestrians \cite{desire, socialgan}. We employ the encoder-decoder model  from Section \ref{sec:enc-dec} for this task, using the Stanford Drone Dataset (SDD) \cite{sdd}. This dataset contains trajectories of, among others, pedestrians, cyclists and skateboarders in eight different scenes. 
Input to the prediction model is a concatenation of two features: 1) relative 2-D coordinates, consisting for each trajectory of the absolute starting point and subsequently x- and y-offsets to the previous point; and 2) simple semantic maps of the environment, processed by a CNN and centered at the agent's location. These semantic maps are binary images of size 64 x 64, in which each pixel describes membership in the classes obstacle and free space.
\section{Results}
\label{sec:results}
In this section we illustrate our extensive experimental evaluation aimed at analyzing the performance of the proposed MHP models. As our goal predominantly is providing a viable general ambiguous prediction framework, for one we compare against non-ambiguous models (SHP), to show the superiority of multiple hypotheses. Further, we provide a competitive comparison of our proposed framework against other well-performing methods yielding multiple hypotheses. Finding best-performing models or outperforming fine-tuned expert ones for specific tasks and datasets is of lesser interest in this work, for the interested reader in the respective sections we compare against some state-of-the art models of this kind nevertheless.

In the field of multi-modal prediction, among the best performing and most commonly used methods are MDNs, MCL and CVAEs, thus we compare against these. In general, we try to use similar models throughout, e.g., an MCL model is identical to the MHP one up to the specific multi-modal part (i.e., using the same input features and exhibiting the same hidden size). As mentioned, other baselines are the SHP models, which are also the same models as presented in Sections \ref{sec:s2s} to \ref{sec:gen}, except missing the MHP parts (i.e. having no duplicate MHP layers, and just outputting one hypothesis). For a fairer comparison, they are extended with the possibility of predicting multiple hypotheses, if possible.

MDNs are only defined for regression problems, and also cannot be applied in a meaningful way for encoder-decoder models. Thus, these are not considered here, but a comparison is done in \cite{Rupprecht2017LearningIA}, yielding favorable results for the MHP framework.
Implementation details about the used models are given in the respective sections.

To better understand differences in computational requirements of the tested algorithms, wall clock times needed for training ($t_t$) and inference ($t_i$) are listed in the following (in $ms$, per mini-batch)\footnote{Note that running times vary based on hardware. All experiments are done using a standard laptop containing an NVIDIA Quadro M2000 GPU.}. For all our experiments, Adam optimizer is used with a learning rate of 0.001. We split the available data into training, validation and test set with a 60-20-20 ratio, employing early stopping on the validation set. All quantitative and qualitative results in the following sections are reported on the test set.

\subsection{Classification}
% In this subsection we show results of the sequence-to-sequence prediction models introduced in Section \ref{sec:s2s}, applied to the toy intersection problem \ref{sec:toyclass} and the problem of predicting lane change \ref{sec:LCP}. 
As simple extensions to a standard SHP model we introduce SHP$^*$, which outputs all classes whose predicted probabilities exceed a certain threshold $\gamma$. 
%Note that setting $\gamma = 0$ equals predicting all possible labels and $\gamma = max$, where $max$ is the maximal predicted probability for each sample, reduces to the standard SHP model. 
For all problems we test $\gamma = 0$ and a problem specific $\gamma$, which returns best results w.r.t. to the M2 metric. The common metrics Precision (Pr), Recall (Re) and the associated F1 score ($F1 = 2 \cdot \frac{Pr \cdot Re}{Pr + Re}$) are used for evaluation, either using an oracle ($Pr_O$ and $Re_O$) or as defined in Section \ref{sec:mm-metrics}.
% For the oracle case the standard definition of $Pr = \frac{1}{\vert M \vert} \sum_{c \in C}\frac{tp_c}{\sum{c' \in C}tp_{c'} + fp_{c'}}$ and $Re = ...$ is used. \todo{maybe introduce prediction in chapter 5, mentioned it is per class precision, and then also define per class MHP precision}.
An MCL model is further used for comparison: It contains $M$ independent copies of the SHP model, during training gradients are only backpropagated to the best hypothesis (which resembles our MHP training with $\epsilon = 0$). For classification problems, we are not aware of meaningful (C)VAE extensions.
\paragraph{Toy Classification}
\label{sec:toyresults}
For all models an LSTM network with 512 hidden units is used, the mini-batch size is 32, the M2 metric parameter $\tau$ is  set to 1, due to the well-behaved structure of the data, and $\epsilon = 0.15$.  The synthetic dataset consists of 10000 trajectories with 75 points each.

\begin{figure}[!t]
\centering
\subfigure[Results of the SHP model.]{
\includegraphics[scale=0.09]{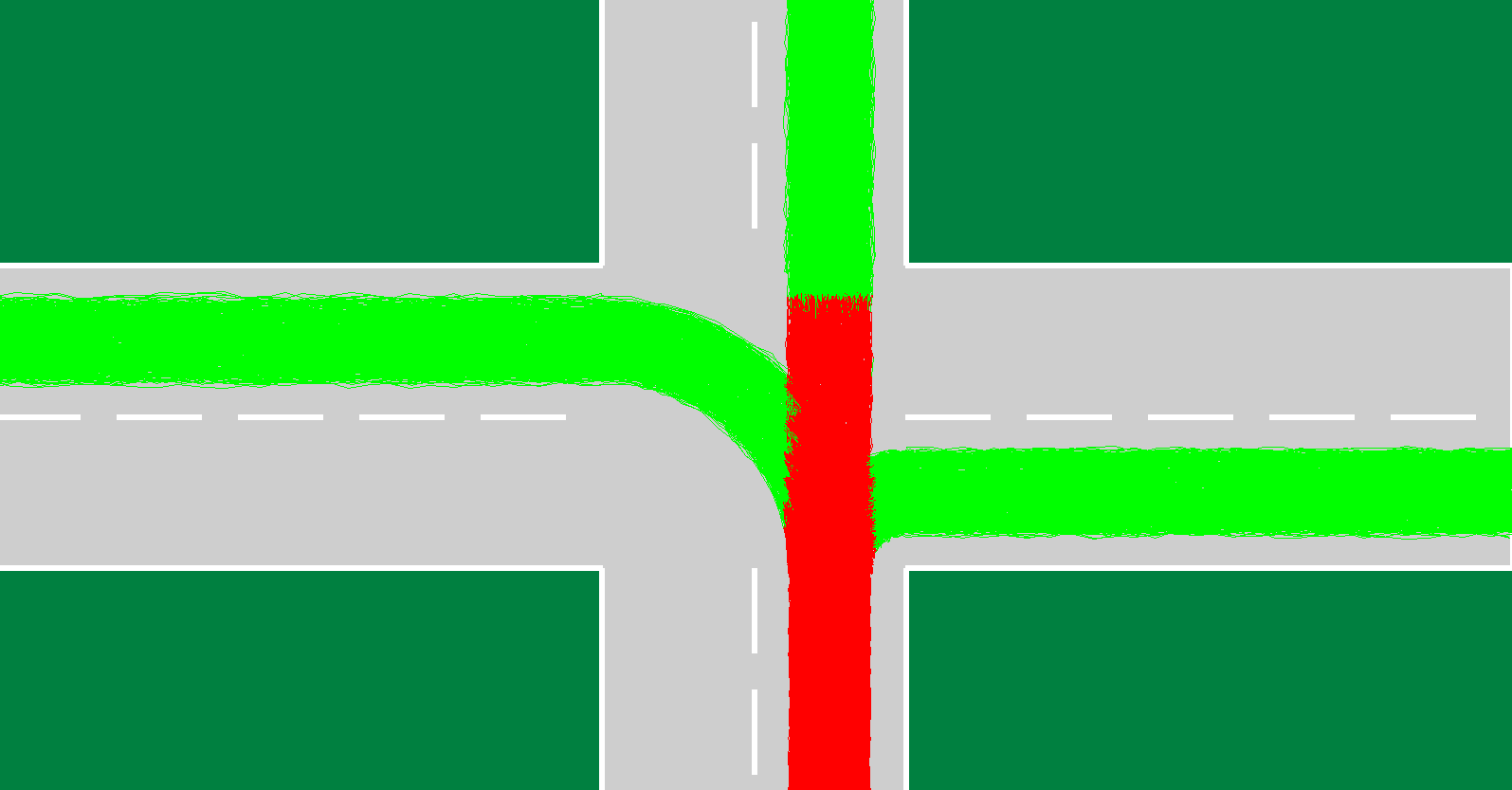}
}
\subfigure[Results of the MHP model.]
{
\includegraphics[scale=0.09]{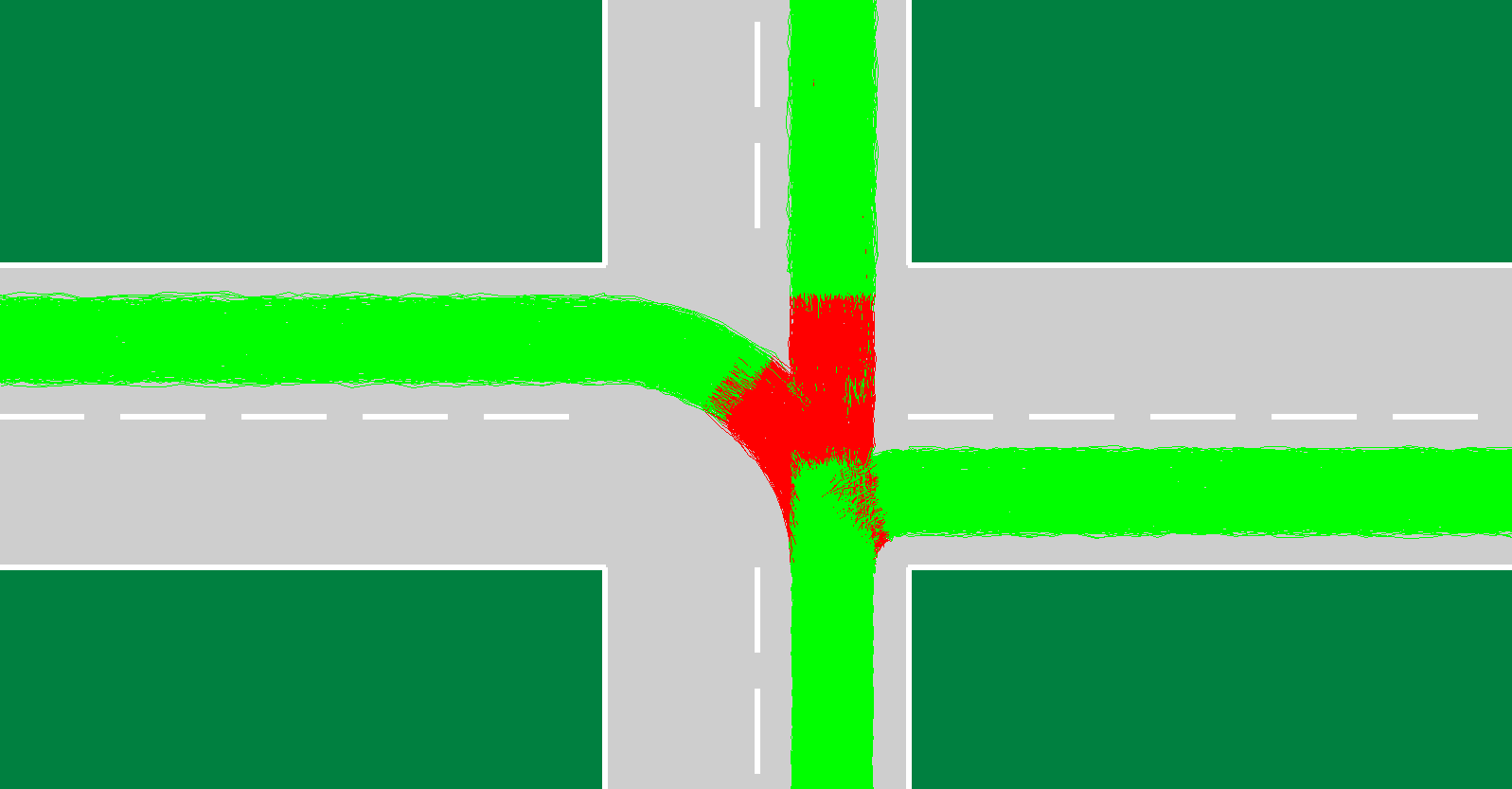}
}
\caption{
Correctness of predictions w.r.t. to the M2 metric: If the set of predictions equals the set of new labels, the point is drawn in green, otherwise in red.}
\label{fig:tc-correct}
\end{figure}

Quantitative results are shown in Table \ref{tab:toyclass}. A standard SHP model picks a random prediction when multiple ones are equally likely, but converges to the correct prediction once the situation becomes unambiguous. This is reflected by an oracle Precision and Recall of roughly $2/3$. The MHP model fares much better with scores of nearly $1$. Setting $\gamma = 0$ reveals a first weakness of the oracle metric, as precision and recall go up to $1$: all 3 existing classes are predicted at each timestep, thus the best hypothesis always equals the ground truth. The M2 metric penalizes this behavior with a lower precision. Conversely, due to the re-labeling, the standard SHP model achieves a near perfect M2 precision (since the predicted hypothesis is nearly always correct), but a low recall. Figure \ref{fig:tc-correct} visualizes these results.
Similar to the SHP model with $\gamma = 0$, the MCL model always outputs 3 different hypotheses. This behaviour is caused by the absence of punishment for (wrongly) diverse predictions due to the principles of MCL ($\epsilon = 0$), and represented by perfect oracle scores, but shows through deductions in the M2 metric.
Since the MCL model consists of $M$ full expert-models, the number of parameters as well as time needed for training and inference grows. In contrast to this, the MHP model only contains slightly more parameters than the SHP one, resulting in near-identical running times.
% i.e. the correctness of the predictions made by the SHP and MHP models w.r.t to the MHP metric. 
% As discussed, the SHP model is wrong in areas with ambigious outcomes, the MHP model only in a small area around the center of the intersection. This could be due to the fact, that these samples are the hardest to classify, but also stemming from artifacts caused by the rectangular shape of the M2 metric polytopes. 
Although some categories are won by other models, overall, the MHP model performs best, proving a correct understanding of the ambiguous nature of the problem.
\begin{table}[!b]
\caption{Results of the toy classification task ($\gamma$ listed in brackets).}
%\vskip 0.15in
\centering
\small
\label{tab:toyclass}
\setlength{\tabcolsep}{4pt}
\begin{tabular}{|c|c|c|c|c|c|c|c|}
 \hline
Name  & $Pr_O$ & $Re_O$ & $Pr_{M2}$ & $Re_{M2}$ & $F1_{M2}$ & $t_t$ &  $t_i$\\
 \hline
SHP & 0.683 & 0.683 & \textbf{0.999} & 0.649 & 0.787 & \textbf{205} & \textbf{105} \\
SHP$^*$ ($0$) & \textbf{1.0} & \textbf{1.0} & 0.676 & \textbf{1.0} & 0.807 & 205 & 105\\
SHP$^*$ ($0.01$) & 0.903 & 0.903 & 0.750 & 0.893 & 0.815 & 205 & 105 \\
MCL & \textbf{1.0} & \textbf{1.0} &0.677 & \textbf{1.0} & 0.807 & 535 & 237 \\
MHP & 0.999 & 0.999 & 0.980 & 0.984 & \textbf{0.982} & 210 & 113 \\
\hline
\end{tabular}
\end{table}
\begin{figure}[!t]
\centering
\subfigure[The three polytopes are depicted for the classification task with discrete labels.]{
\includegraphics[scale=0.09]{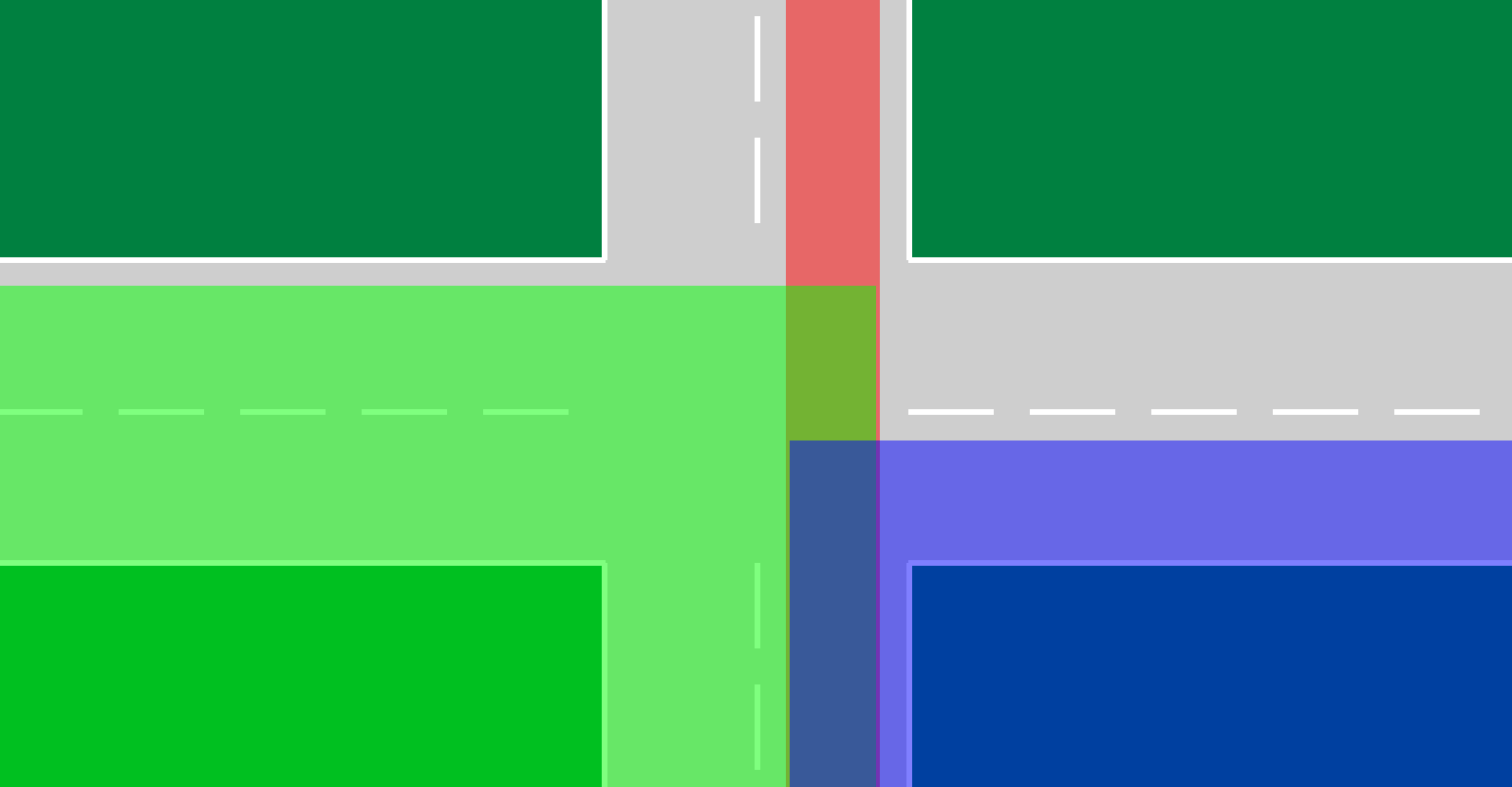}
}
\subfigure[The clustering results for the prediction task with continuous labels are drawn, the cluster centers are depicted in yellow.]
{
\includegraphics[scale=0.09]{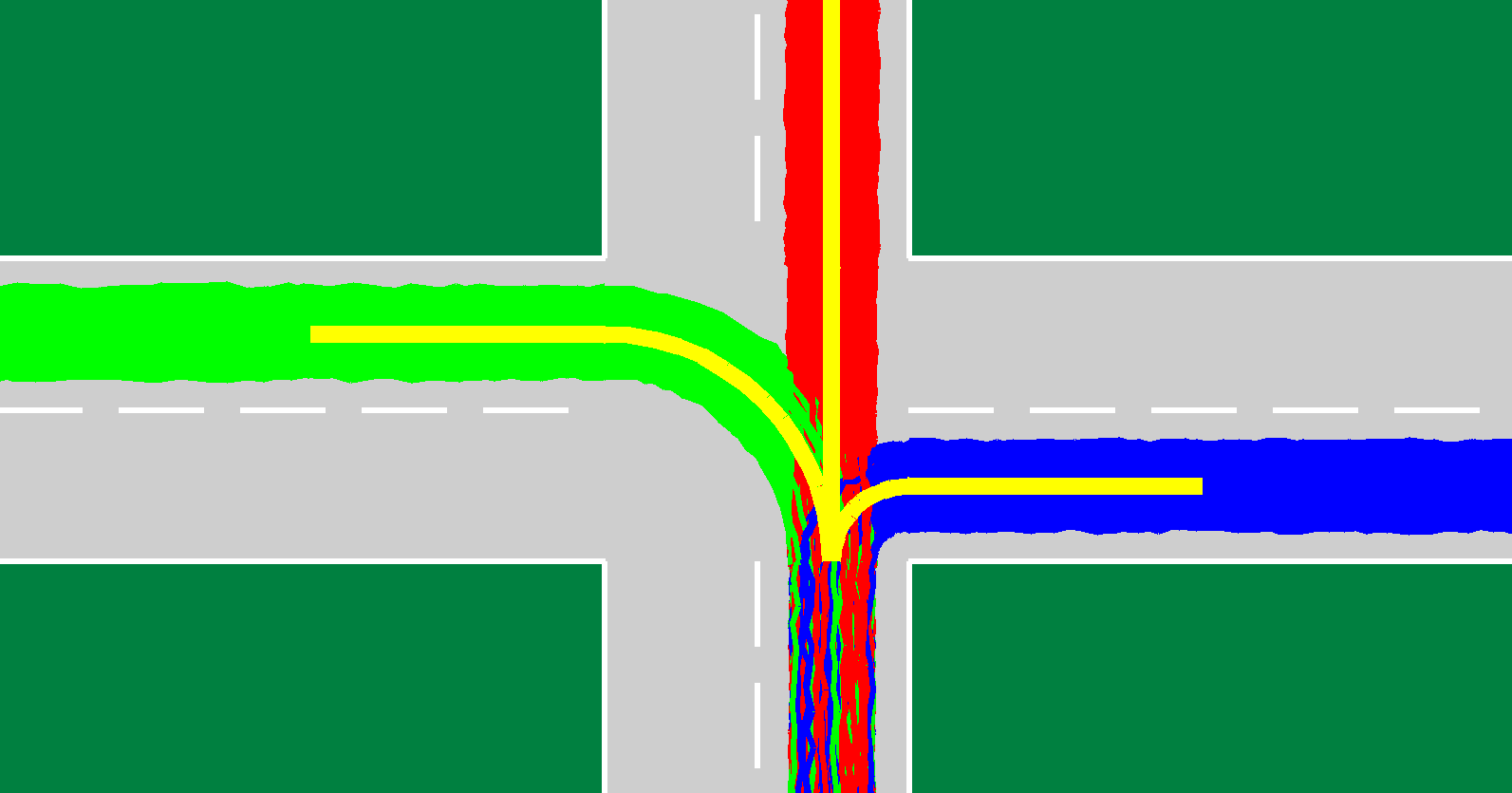}
}
\caption{Visualizations of re-labelling step for M2 metric.}
\label{fig:m2metric}
\end{figure}
Figure \ref{fig:m2metric}a shows the polytope obtained by applying the re-labelling procedure of the M2 metric for this task.
\paragraph{Lane Change Prediction}
%As mentioned in Section \ref{sec:mm-metrics}, parameter $\tau$ should be chosen such that the resulting labeling resembles our human understanding of ambiguity in the problem. We find $\tau = 0.85$ to be well suited for this.
We set $\tau = 0.85$, so that the resulting labelling resembles our human understanding of ambiguity in the problem.
%(compare Figure \ref{fig:m2sample}).
%shows two samples and their respective new labels, which coincide with human labeling. From the NGSIM dataset we use the subsets US 101 and I80, which contain about 1.200.000 single frames. -> move to supplementary
% Table \ref{tab:ngsim} shows the number of samples of each possible class before and after the relabeling.
% \begin{table}[!b]
% \small
% \label{tab:ngsim}
% \begin{tabular}{|c|c|}
% \hline
% Label & \# Samples \\
% \hline
% L / F / R (original) & 13241 / 1179384 / 3636 \\
% \hline
% LFR & 1403 \\
% LF / RF /  LR & 94358 / 179085 / 178 \\
% L / F / R & 12590 / 906532 / 3447 \\
% \hline
% \end{tabular}
% \caption{Number of samples, ordered by label class.}
% \end{table}
For all models an LSTM size of 128 is used, a mini-batch size of $256$ and $\epsilon = 0.15$. 
The results can be found in Table \ref{tab:lcp} and are similar to those discussed in the previous paragraph, the MHP model performing best.
Note that our used SHP model is the same as in \cite{lcpattention}, thus a direction comparison to a state-of-the art (non-ambigious) model is given. To the best of our knowledge, no ambiguous models for this kind of task are employed.
\begin{table}[!b]
\caption{Results on the NGSIM dataset ($\gamma$ listed in brackets).}
%\vskip 0.15in
\centering
\small
\label{tab:lcp}
\setlength{\tabcolsep}{4pt}
\begin{tabular}{|c|c|c|c|c|c|c|c|}
\hline
 Name & $Pr_O$ & $Re_O$ & $Pr_{M2}$ & $Re_{M2}$ & $F1_{M2}$ & $t_t$ &  $t_i$ \\
 \hline
SHP & 0.766 & 0.766 & \textbf{0.797} & 0.747 & 0.771 & \textbf{388} & \textbf{263} \\
SHP$^*$ (0) & \textbf{0.952} &\textbf{0.952} & 0.352 & \textbf{0.952} & 0.514 & 388 & 263\\
SHP$^*$ (0.55) & 0.799 & 0.799 & 0.572 & 0.779 & 0.660 & 388 & 263\\
MCL & \textbf{0.952} &\textbf{0.952} & 0.364 & \textbf{0.952} & 0.527 & 1005 & 637\\
MHP & 0.905 & 0.905 & 0.732 & 0.906 & \textbf{0.810} & 658 & 506\\
\hline
\end{tabular}
\end{table}

% \begin{figure}[!t]
% \centering
% \subfigure[A lane change to the left is executed smoothly, the sample is labeled solely with L.]{
% \includegraphics[scale=0.35]{images/L.png}
% }
% \subfigure[This sample is labeled with LF, also a human might be unsure about the driver's intentions.]
% {
% \includegraphics[scale=0.35]{images/F.png}
% }
% \caption{
% Depiction of the new labels generated by the M2 metric.}
% \label{fig:m2sample}
% \end{figure} -> supllementary

\subsection{Prediction}
% In this section the results of encoder-decoder models are shown, applied to a toy and a real problem. 
A standard SHP encoder-decoder model cannot easily be extended to output multiple predictions, thus no extension of this is analyzed.
We analyze MCL and CVAE extensions though: Again, the MCL model consists of $M$ copies of the encoder-decoder SHP. In the CVAE, the encoder produces a 20-dimensional latent vector $z$, represented by mean and standard deviation. Via the reparametrization trick the decoder samples from this to generate $M$ hypotheses. Similar to \cite{desire}, the average error over all hypotheses is used, in addition to the Kullback-Leibler divergence between $z$'s distribution and a unit normal distribution.
To evaluate the outcomes we use the  Final Displacement Error (FDE) and Average Displacement Error (ADE) metrics, which denote the metric distance of the prediction to the ground truth considering either only the last timestep or all, respectively. 
% For the oracle version the best prediction is used, the MHP extension is described in Section \ref{sec:mm-metrics}.
For the clustering used in the M2 metric, we convert trajectories consisting of $n$ timesteps to a $2n$-dimensional space, and in this employ mean-shift clustering.

\paragraph{Toy Prediction}
% We first show some results on a toy example to better understand the behaviour of our models. 
Again, we use the toy intersection from Section \ref{sec:helpertoy}, on which we simulate 10000 trajectories containing 60 timepoints each. The first 30 are fed to the encoder, and the decoder is expected to produce the remaining trajectory.
Both LSTMs consist of 64 hidden units, $\epsilon = 0.15$, $\tau = 1$ and mini-batch size is 256. The bandwidth used for mean-shift clustering is determined automatically by the algorithm. 

\begin{figure}[!t]
\centering
\subfigure[Prediction of the SHP model.]{
\includegraphics[scale=0.09]{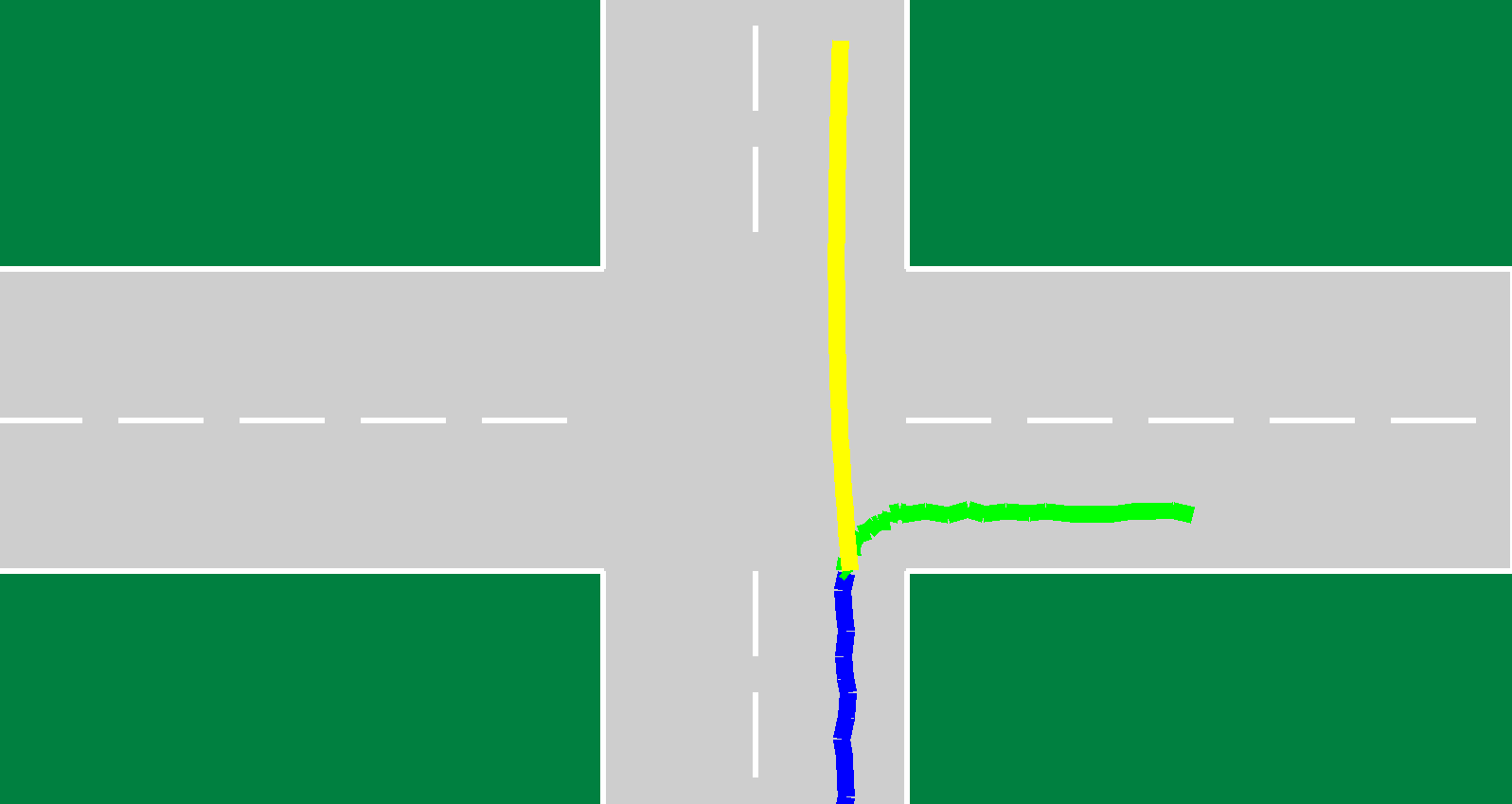}
}
\subfigure[Prediction of the MHP model.]
{
\includegraphics[scale=0.09]{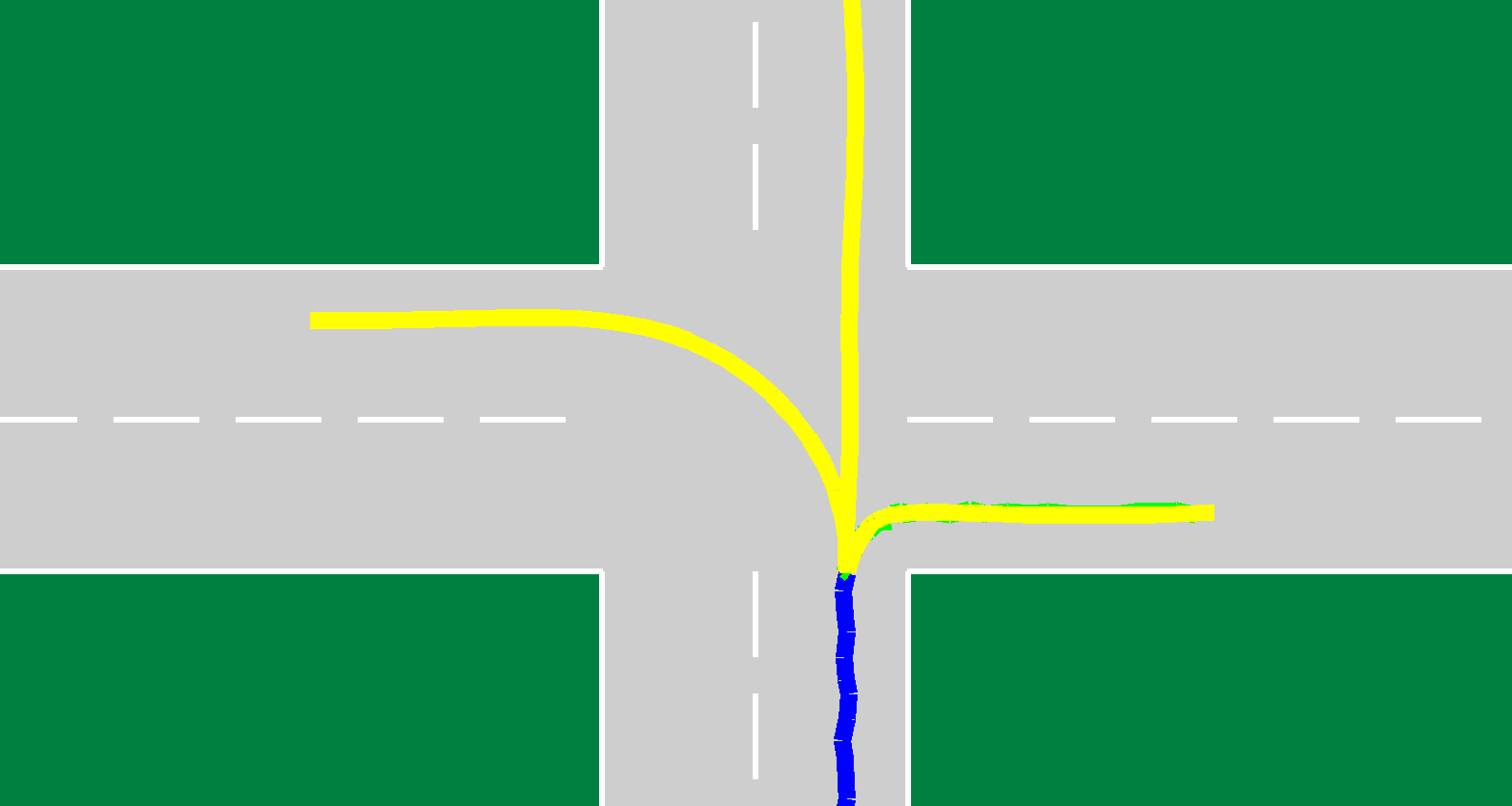}
}
\caption{Resulting prediction for a sample trajectory. The input to the decoder is drawn in blue, the ground truth in green, and the predictions in yellow.
}
\label{fig:toypredresults}
\end{figure}

The predictions for one sample can be seen in Figure \ref{fig:toypredresults}, for more we refer to the supplementary video. Whereas the SHP model predicts one random path, the MHP model nicely accounts for all possible outcomes. The corresponding quantitative results over the test dataset are reported in Table \ref{tab:toypred}.
The SHP model only outputs one mean prediction, resulting in bad overall scores. Also the CVAE model shows problems generating diverse outputs, falling back to the mean as well. This is to be expected, as inputs are very similar, resulting in an average distribution. Both the MCL and MHP model perform much better. Again, the MCL model exhibits a longer training and inference time though, while CVAE and MHP perform similarly timewise.
Figure \ref{fig:m2metric}b shows the resulting clusters used for the M2 metric.

\begin{table}[!b]
\caption{Results of the toy prediction task.}
%\vskip 0.15in
\centering
\small
\label{tab:toypred}
\setlength{\tabcolsep}{4pt}
\begin{tabular}{|c|c|c|c|c|c|c|}
\hline
 Name & $FDE_O$ & $ADE_O$ & $FDE_{M2}$ & $ADE_{M2}$ & $t_t$ & $t_i$\\
\hline
SHP & 2.82 & 4.60 & 2.29 & 3.64 & \textbf{112} & \textbf{49} \\
MCL & \textbf{0.15} & \textbf{0.16} & 0.53 & \textbf{0.53} & 320 & 147\\
CVAE & 2.76 & 4.53 & 1.99 & 2.95 & 218 & 83 \\
MHP & 0.26 & 0.22 & \textbf{0.51} & \textbf{0.53} & 210 & 88\\
\hline
\end{tabular}
\end{table}

\paragraph{Trajectory Prediction}
Using the SDD dataset we experiment with 3 and 5 hypotheses, in addition to the SHP encoder-decoder. For all models we use an LSTM size of 256 and a sequence length of 8 (3.2s) for the encoder and 12 (4.8s) for the decoder, to comply with existing works in this field.
Further, $\epsilon = 0.15$, $\tau = 1$ and mini-batch size 256. For the mean-shift clustering we use a bandwidth of 250. 

\begin{figure}[!t]
\centering
\subfigure[Prediction using 5 hypotheses.]{
\includegraphics[scale=0.16]{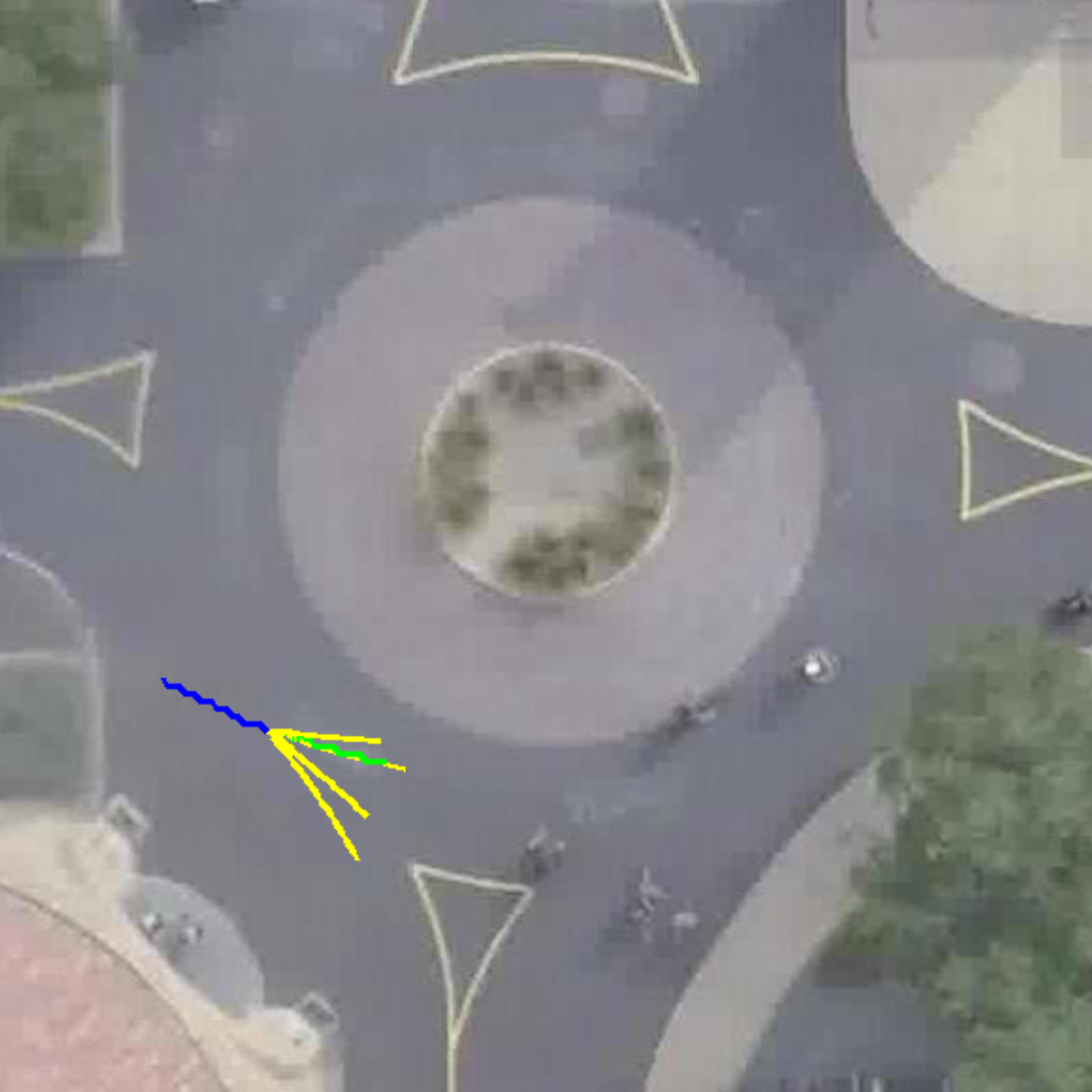}
}
\subfigure[Prediction using 3 hypotheses.]
{
\includegraphics[scale=0.16]{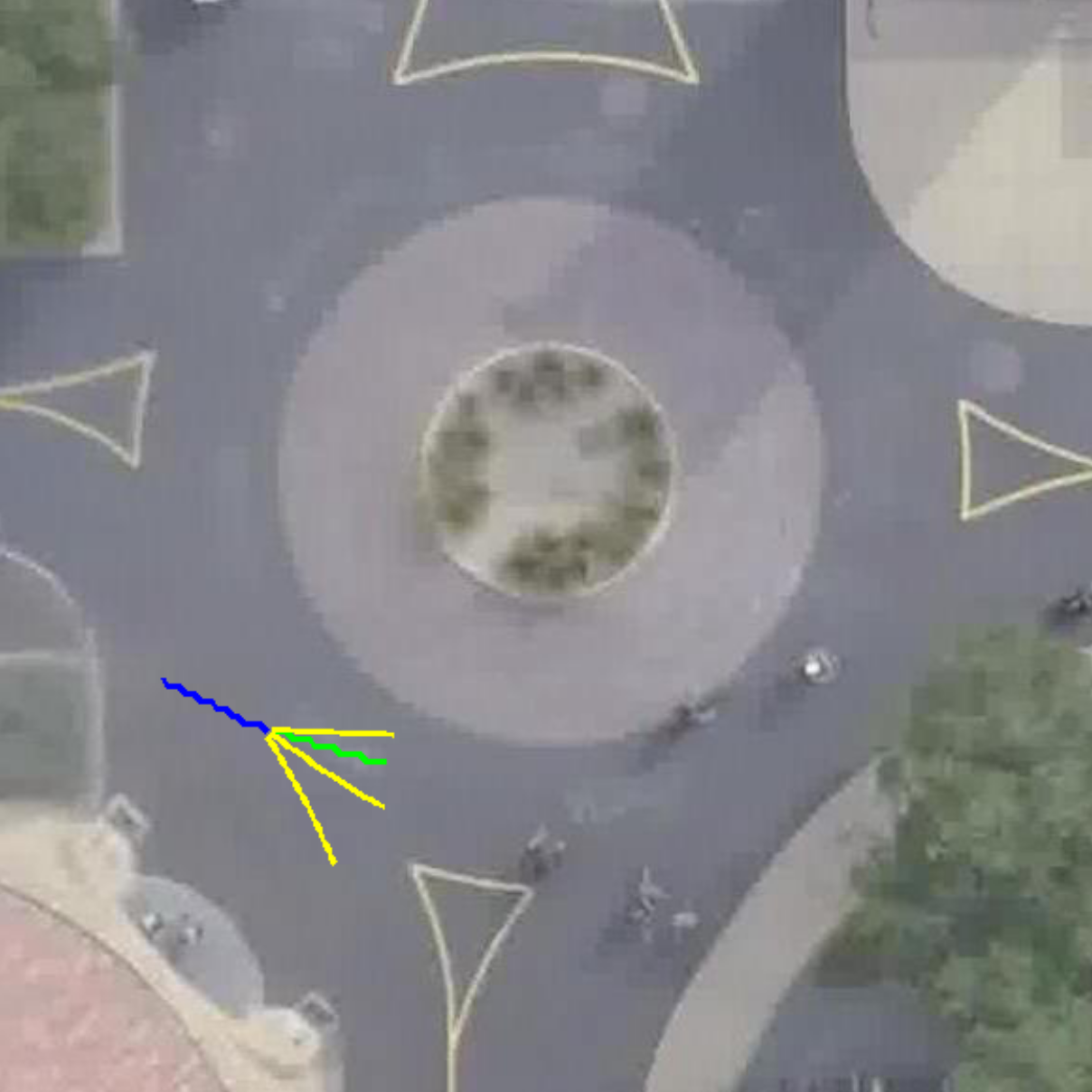}
}
\subfigure[Prediction of the SHP model.]
{
\includegraphics[scale=0.16]{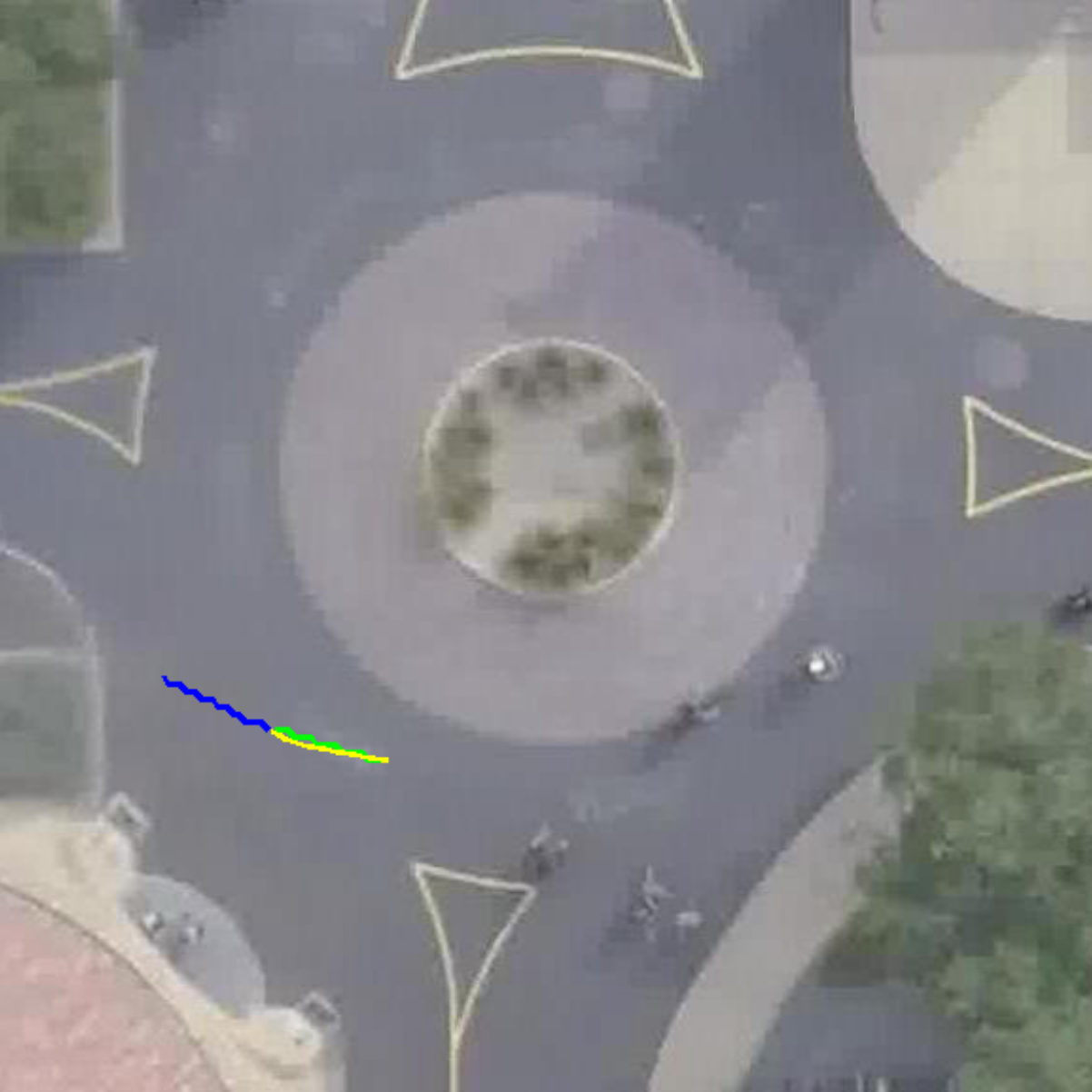}
}
\caption{
Resulting predictions in an ambiguous scene of the SDD.}
\label{fig:resultsonsdd}
\end{figure}

Figure \ref{fig:resultsonsdd} shows the results on the ambiguous scene, which was hinted in Figure \ref{fig:teaser}. The SHP model simply predicts a linear follow-up trajectory, whereas the MHP models cover the full space of possible directions, accurately matching the existing street topology. More hypotheses provide more fine-grained predictions. Again we refer to the supplementary video for more results, also including those of other models.
%In Figure \ref{fig:sddresultsb} more linear motions are considered, the agents mostly following the existing sidewalk. All predictions thus lie in a narrow space, predicting a straight trajectory. Note, though, that when employing 5 hypotheses, there is one which ends nearly immediately. This could represent a rare but plausible outcome of a person stopping, e.g. to chat. 
Table \ref{tab:sddres} shows quantitative results over the full test set, proving that the MHP models perform better than the SHP one. 
The VAE models improve over the toy task given the more diverse input, still sometimes lack diversity and miss hypotheses, as shown by higher oracle scores.
Overall, regarding the oracle metric the MHP models perform best, w.r.t. the M2 metric the performances of the ambiguous models are close together - all seem to show an understanding of multimodality when faced with this problem.
As before, the MCL models exhibit the longest running times, while VAE and MHP are close together, with MHP scaling slightly better with growing $M$.
Further we compare our methods to SoPhie GAN \cite{sophie} and the DESIRE framework \cite{desire} (results for this obtained from the interpolation in \cite{sophie}), for a reference to state-of-the art models. For our experiments we use the SDD part of the TrajNet dataset \cite{sadeghiankosaraju2018trajnet}, in which the sequences of the SDD have been preformatted to the desired sequence lengths and overly linear trajectories been filtered out. As labels for the test set are not public (yet) and multiple hypotheses not supported in the score calculation, we created our own train / test split from the public training set, which might explain the slightly better scores of our models.

%For the oracle metric a significant improvement can be seen, for the M2 metric this is less. 
% One reason for this is, that like with the models, we use simple and general procedures for calculating this (i.e. the clustering function) for demonstration purposes. There might be better, problem specific solutions. 
%This is due to the label space being vast and relatively unstructured, as the agents walk all over the space, resulting in multiple clusters. In more structured scenarios, like cars following lanes, the results differ. -> move to supplementary

% \begin{figure}[!t]
% \centering
% \subfigure[Prediction using 5 hypotheses.]{
% \includegraphics[scale=0.3]{images/hyang-5-small.png}
% }
% \subfigure[Prediction using 3 hypotheses.]
% {
% \includegraphics[scale=0.3]{images/hyang-3-small.png}
% }
% \subfigure[Prediction of the SHP model.]
% {
% \includegraphics[scale=0.3]{images/hyang-1-small.png}
% }
% \caption{
% Resulting predictions in a non-ambiguous scene of the SDD.}
% \label{fig:sddresultsb}
% \end{figure}

\begin{table}[!b]
\caption{Results on the SDD.}
%\vskip 0.15in
\centering
\small
\label{tab:sddres}
\setlength{\tabcolsep}{4pt}
\begin{tabular}{|c|c|c|c|c|c|c|}
 \hline
Name  & $FDE_O$ & $ADE_O$ & $FDE_{M2}$ & $ADE_{M2}$ & $t_t$ & $t_i$\\
\hline
SHP & 35.94 & 17.76 & 49.14 & 40.36 & \textbf{238} & \textbf{143} \\
MCL 3 & 27.52 & 14.46 & 45.02 & 41.89  & 338 & 187\\
MCL 5 & 25.31 & 13.58 & 43.81 & 42.52  & 450 & 227\\
VAE 3 & 35.07 & 17.16 & 44.06 & 38.86  & 311 & 166 \\
VAE 5 & 34.06 & 16.62 & \textbf{41.86} & \textbf{38.23}  & 377 & 193\\
MHP 3 & 27.20 & 14.36 & 42.66 & 39.03  &  312 & 175 \\
MHP 5 & \textbf{24.89} & \textbf{13.48} & 41.96 & 39.11  & 355 & 187\\ 
DESIRE & 34.05 & 19.25 & - & -  & - & - \\
SoPhie & 29.38 & 16.27 & - & -  & - & - \\ 
\hline
\end{tabular}
\end{table}

\subsection{Sequence Generation}
In this section we demonstrate our MHP extension for sequence generation models, predicting trajectories on the toy intersection. We again use the FDE and ADE metrics.
The synthetic dataset consists of 10000 trajectories with 25 points each. In sequence generation scenarios, a standard SHP model can only be modified in a very specific way, if at all, to output multiple hypotheses. Experiments did not yield any usable results for SHP extensions. Further, CVAE or MCL models cannot be extended to this application scheme in a trivial way.
Thus here we only compare a standard SHP model to the MHP extension.
Both LSTMs used have a hidden size of 256, same for mini-batch size, $\epsilon = 0.15$ and $\tau = 1$. 

\begin{figure}[!t]
\centering
\subfigure[Result of the SHP model.]{
\includegraphics[scale=0.09]{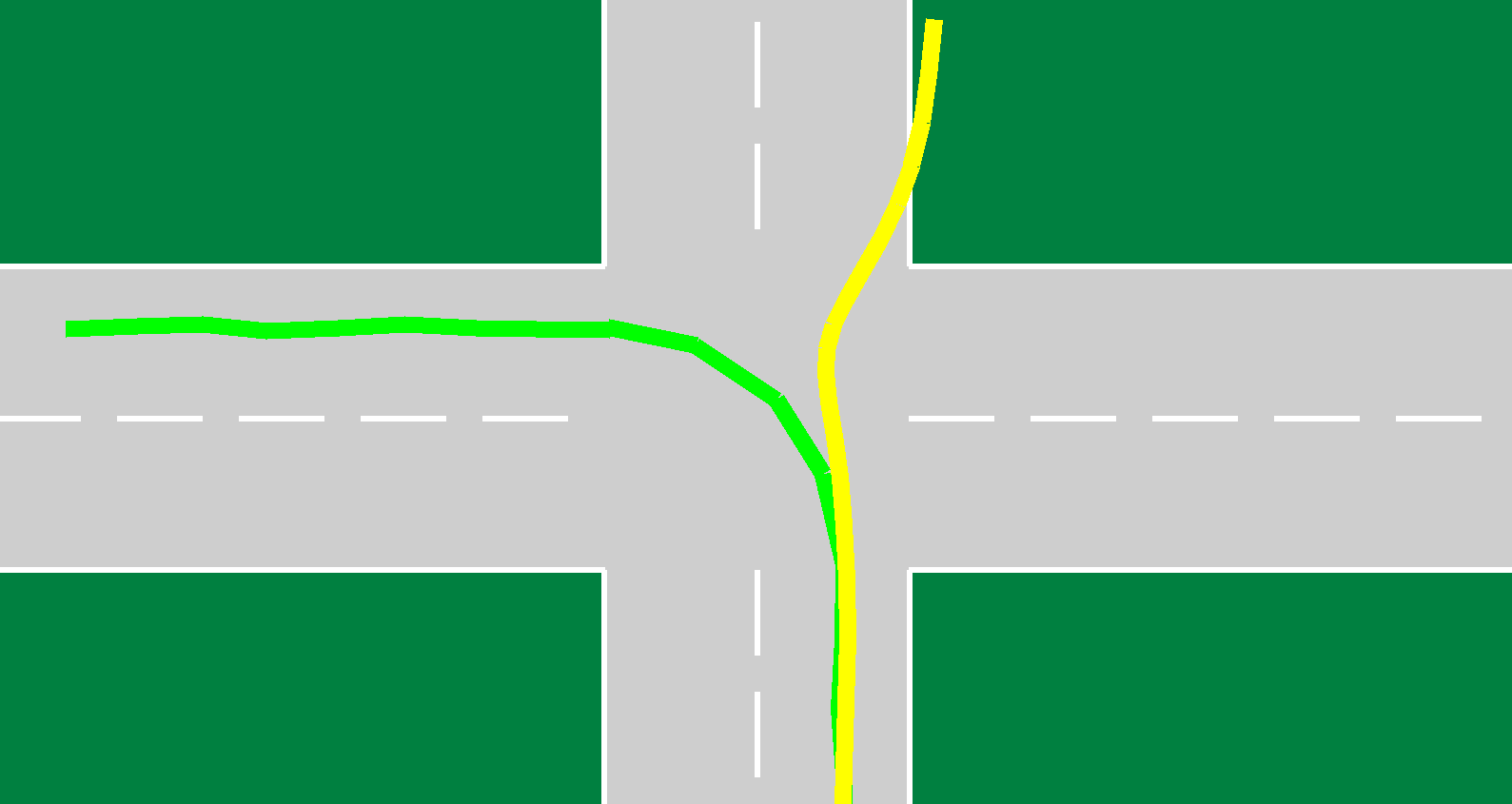}
}
\subfigure[Result of the MHP model.]
{
\includegraphics[scale=0.09]{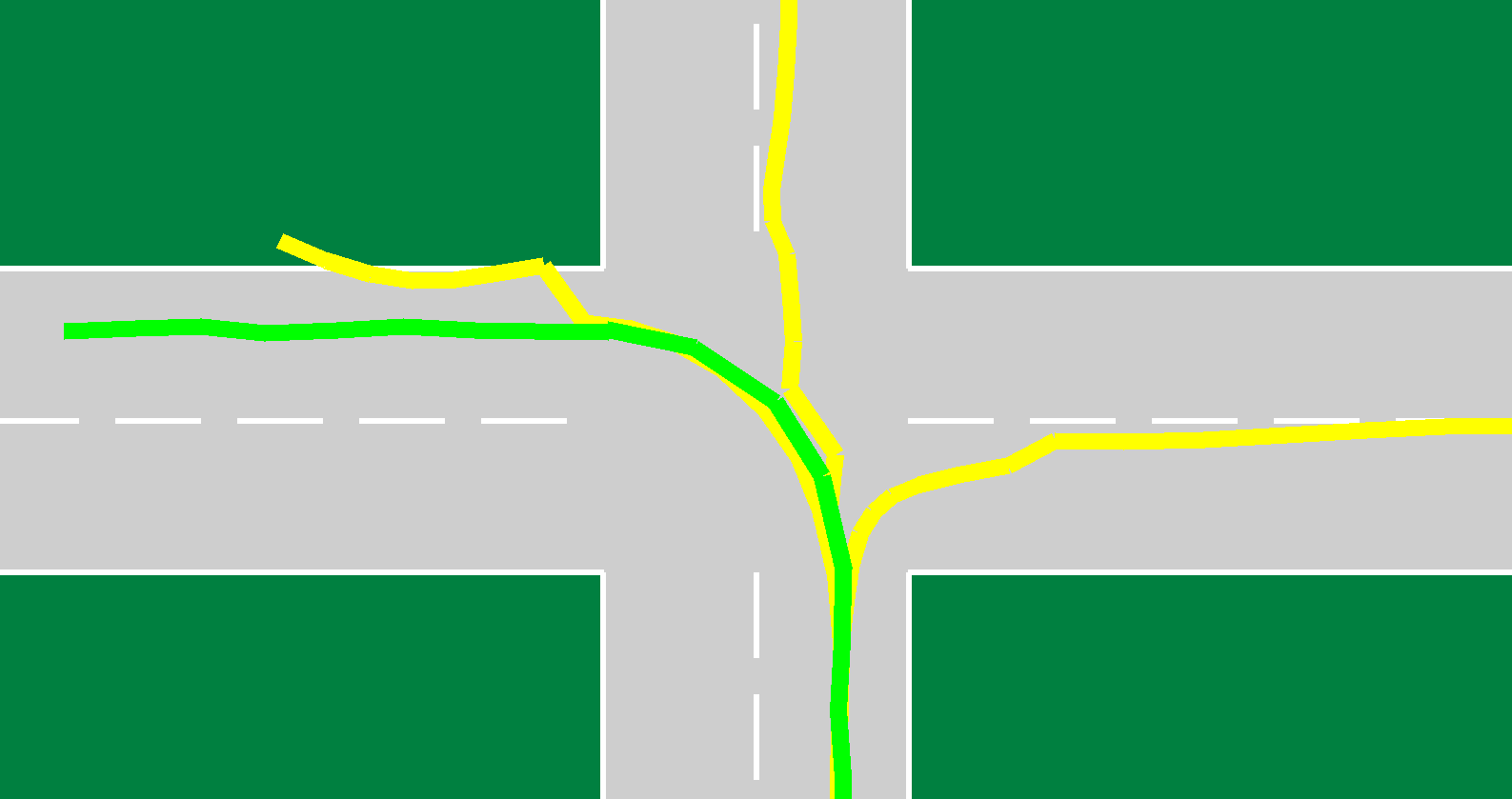}
}
\caption{
Prediction results on one sample trajectory. The ground truth is depicted in green, the predictions in yellow.}
\label{fig:autoresults}
\end{figure}

After training, for each test trajectory we initialize the model with 5 points and infer 20. Figure \ref{fig:autoresults} shows the resulting predictions for one sample, in Table \ref{tab:auto} the results for the test set are listed. Again, the MHP model outperforms the SHP model. 
% In Figure \ref{fig:tree} two trees created at different timepoints, one early and one right at the intersection, are shown. In the first case, the resulting tree is very narrow, as there is no ambiguity in the predictions. In the second case the tree correctly covers the full possible label space of the intersection, and the function $\texttt{CheckSplit}$ returns true. 
While training times of both models are similar, inference equals a single forward pass in the SHP model, while in the MHP model a tree is created to generate multiple hypotheses, resulting in higher inference times.

\begin{table}[!b]
\caption{Results on the toy sequence generation task.}
%\vskip 0.15in
\centering
\small
\label{tab:auto}
\setlength{\tabcolsep}{4pt}
\begin{tabular}{|c|c|c|c|c|c|c|}
\hline
Name & $FDE_O$ & $ADE_O$ & $FDE_{M2}$ & $ADE_{M2}$ & $t_t$ & $t_i$\\
\hline
SHP & 18.26 & 5.07 & 14.07 & 4.13 & \textbf{59} & \textbf{188} \\
MHP & \textbf{5.45} & \textbf{1.99} & \textbf{5.48} & \textbf{2.06} & 67 & 24433\\
\hline
\end{tabular}
\end{table}

% \begin{figure}[!t]
% \centering
% \subfigure[Tree at step 1.]{
% \includegraphics[scale=0.09]{images/tree-1.png}
% }
% \subfigure[Tree at the beginning of the intersection.]
% {
% \includegraphics[scale=0.09]{images/tree.png}
% }
% \caption{
% Visualization of the created tree at different timepoints. Points of the tree are drawn in orange, the resulting hypotheses produced by \texttt{ChooseTreePaths}, in case \texttt{CheckSplit} returns true, in yellow.}
% \label{fig:tree}
% \end{figure}
\section{Conclusion}
In this paper we have proposed an extension of the Multiple Hypothesis Prediction model, resulting in a  universally applicable framework for dealing with ambiguity and uncertainty in sequential problems. Experiments with different architectures and in fields as diverse as trajectory analysis and maneuver prediction show the wide applicability and superior performance of ambiguous models over standard non-ambiguous ones in the considered cases. 
Additionally, as opposed to other multi-modal prediction models, the MHP extension proves to be applicable for all kinds of problems and is among the best performing methods for each of them, while simultaneously exhibiting minimal parameter overhead.
Furthermore, to better assess upon problems featuring data with multiple possible labels, we have introduced a novel metric, which we posit be considered whenever dealing with ambiguous problems. Across diverse problems and tasks we showed its applicability and pitfalls of standard oracle metrics.
\section{Acknowledgements}
We would like to thank Christian Rupprecht and Iro Laina for valuable discussions and feedback.
{\small
\bibliographystyle{IEEEtran}
\bibliography{egbib}
}

\end{document}